\documentclass{article}

\usepackage{microtype}
\usepackage{graphicx}
\usepackage{subfigure}
\usepackage{booktabs} 
\usepackage{amsmath}
\usepackage{amssymb}
\usepackage{amsthm}
\usepackage{textcomp}
\newtheorem{proposition}{Proposition}
\usepackage{algorithm}
\usepackage{algorithmic}
\usepackage[table]{xcolor}
\usepackage{threeparttable}

\usepackage{hyperref}

\usepackage[accepted]{icml2025}

\usepackage{mathtools}

\usepackage[capitalize,noabbrev]{cleveref}

\theoremstyle{plain}

\theoremstyle{definition}

\theoremstyle{remark}

\usepackage[textsize=tiny]{todonotes}

\icmltitlerunning{LMDM:Latent Molecular Diffusion Model For 3D Molecule Generation}

\begin{document}

\twocolumn[
\icmltitle{LMDM:Latent Molecular Diffusion Model For 3D Molecule Generation}



\icmlsetsymbol{equal}{*}

\begin{icmlauthorlist}
\icmlauthor{Xiang Chen}{}
\end{icmlauthorlist}

\icmlcorrespondingauthor{Xiang Chen}{xiangchen981024@gmail.com}

\icmlkeywords{Machine Learning, ICML}

\vskip 0.3in
]



\printAffiliationsAndNotice{\icmlEqualContribution} 

\begin{abstract}
In this work, we propose a latent molecular diffusion model that can make the generated 3D molecules rich in diversity and maintain rich geometric features. The model captures the information of the forces and local constraints between atoms so that the generated molecules can maintain Euclidean transformation and high level of effectiveness and diversity. We also use the lower-rank manifold advantage of the latent variables of the latent model to fuse the information of the forces between atoms to better maintain the geometric equivariant properties of the molecules. Because there is no need to perform information fusion encoding in stages like traditional encoders and decoders, this reduces the amount of calculation in the back-propagation process. The model keeps the forces and local constraints of particle bonds in the latent variable space, reducing the impact of underfitting on the surface of the network on the large position drift of the particle geometry, so that our model can converge earlier. We introduce a distribution control variable in each backward step to strengthen exploration and improve the diversity of generation. In the experiment, the quality of the samples we generated and the convergence speed of the model have been significantly improved.
\end{abstract}
\section{Introduction}
In the field of 3D molecular generation, we expect to obtain richer particle information, in which the interatomic forces and geometric feature representation information play a vital role in maintaining the stability and richness of molecular geometric structure. In the research process, we generally represent the geometric structure in the form of Cartesian coordinates, the molecule as a 3D atomic graph\cite{schutt2017schnet}, the protein as a neighboring spatial graph on amino acids\cite{jing2021learningproteinstructuregeometric}, and use a dual equivariant fractional network to fuse the relative distance and the force between particles into particle score information\cite{satorras2022enequivariantgraphneural}. Therefore, the geometric feature generation model that simulates the covalent bonds formed by intermolecular forces at close range and the van der Waals forces at long distances has great potential for accelerating new drug discovery, catalyst design, and materials science\cite{pereira2016boosting,oord2016wavenet,townshend2021atomd}. After Alphafold's success in protein folding prediction\cite{jumper2021highly}, more and more researchers are developing deep learning models to analyze or synthesize 3D molecules\cite{simonovsky2018graphvae,gebauer2020symmetryadaptedgeneration3dpoint,klicpera2020dimenet}.

They have all made progress in effectiveness. Nevertheless, the diffusion-based 3D generative model still has two non-negligible shortcomings: First, unlike the chemical bonds of 2D generated molecules represented as graphic edges, the geometric shapes of 3D generated molecules are represented as point clouds\cite{satorras2021enflow,gebauer2019symmetry,hoogeboom2022equivariant}. Therefore, there is no clear indication of chemical bonds when generating 3D molecules, which makes it difficult for both EDM and GeoLDM to capture the rich local constraint relationships between neighboring atoms. This defect leads to unsatisfactory training results on large molecular datasets, such as the GEOMDrugs dataset\cite{axelrod2022geom}. The diffusion model moves along the data density gradient at each time step. Therefore, the generation dynamics of a given fixed initialization noise may also be concentrated around common trajectories, resulting in similar generation results, which greatly reduces the diversity of generated molecules.

\begin{figure*}[!t]
    \centering
    \includegraphics[height=0.31\textwidth]{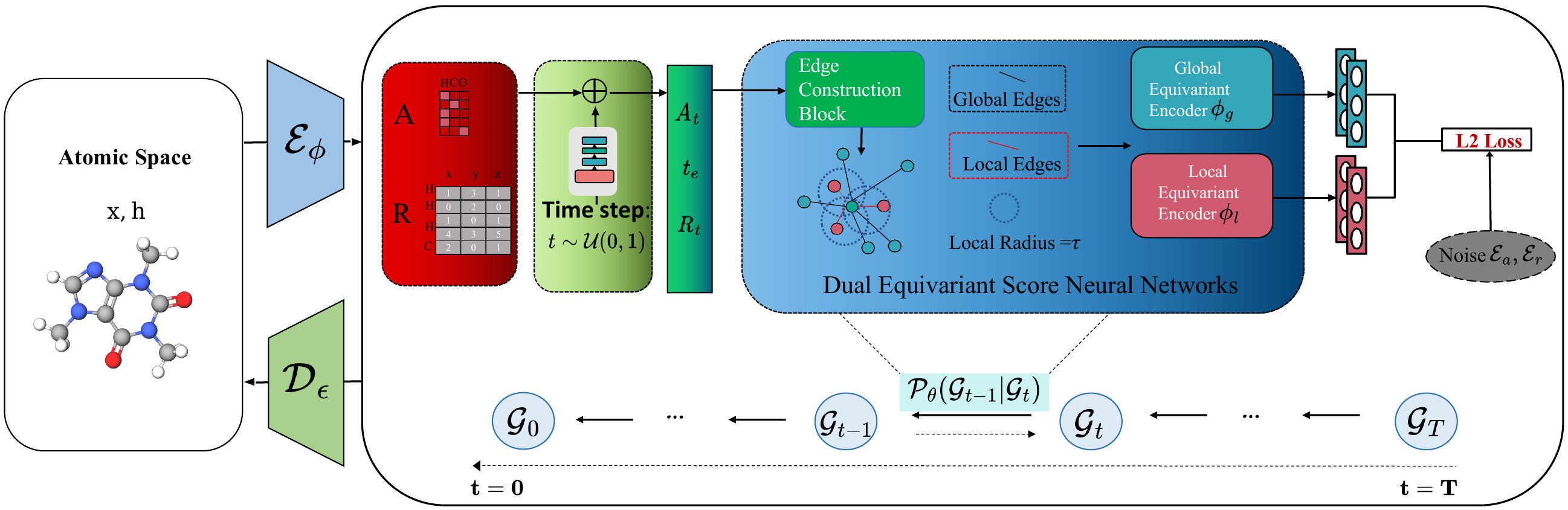} 
    \vspace{-5pt}
    \caption{Illustration of LMDM.
    We outline the training process of the proposed LMDM model. The encoder $\mathcal{E}_{\phi}$ coordinates $\mathrm{x}$ and molecular features $\mathrm{h}$ are encoded into equivariant latent variables $\mathrm{R}$,$\mathrm{A}$, and the time step encoding is used to incorporate the sequential information into the molecular information. We gradually add noise through the latent diffusion transformation $q(\mathcal{G}_{t} \mid \mathcal{G}_{t-1})$ until the latent variable distribution converges to a Gaussian distribution. Similarly, for the reverse generation process, the initial state $\mathcal{G}_{T} \sim \mathcal{N}(0, I)$ is gradually denoised by using the Markov kernel $\mathrm{p}_{\theta}(\mathcal{G}_{t-1} \mid \mathcal{G}_{t})$ and gradually refined by the equivariant denoising dynamics $\mathbf{\epsilon_\theta}(\mathcal{G}_{t}, \mathbf{t})$. The final latent variables $\mathrm{R}$, $\mathrm{A}$ are further decoded by the decoder $\mathcal{D}_{\epsilon}$ to generate the molecular point cloud.
    }
    \label{fig:LMDM}
    \vspace{-10pt}
\end{figure*}
In addition, we also map the features to a regularized latent space, maintaining the key 3D rotation and translation equivariance constraints, which will model a smoother distribution, reduce the difficulty of directly modeling complex structures, and play a role in encoding geometric features, making the model more expressive. At the same time, the latent diffusion model can better control the generation process, which is a promising result in text-guided image generation\cite{rombach2022high}. This enables users to generate specific types of molecules with desired properties. Our model can be extended to many downstream tasks, such as targeted drug design\cite{lin2022diffbp} and antigen-specific antibody generation\cite{luo2022antigenspecific}.

In experiments, we show that our proposed LMDM outperforms the state-of-the-art models EDM and GeoLDM on two molecular datasets (\textit{i.e.}, QM9\cite{ramakrishnan2014quantum}) and GEOM-Drugs~\cite{axelrod2022geom}), especially on the drug-like GEOM-Drugs dataset, which has a large number of atoms in the molecular composition (46 atoms on average, compared to 18 atoms in QM9). We improve the effectiveness and diversity of characterizing the generated molecules by 4.8\% and 30.2\%, respectively, without compromising the effectiveness and stability of the generated molecules in conditional generation.
\section{Relate work}
\textbf{Latent Molecular Diffusion Models.} In order to improve the modeling ability of the model, predecessors have conducted a lot of research, and the model can learn stronger expressiveness in the latent space\cite{dai2018diagnosing,yu2022vectorquantized}. \textit{e.g.}, VQ-VAE\cite{razavi2019generating} proposed discretized latent variables and learned the prior distribution of images through autoregression. \cite{ma2019flowseq} uses a flow-based model as a latent prior and applies it to non-autoregressive text generation. Superior to the simple Gaussian prior that cannot accurately match the encoded posterior, inspired by the variational autoencoder (VAE), \cite{dai2018diagnosing,aneja2021contrastive} proposed to use VAE and energy-based models to learn the latent distribution respectively. These works seem to indicate that by combining the diffusion model with VAE, the model can learn a more accurate latent distribution. In recent years, latent diffusion models have been widely used in various fields and have achieved encouraging results, such as image\cite{vahdat2021score}, point cloud\cite{zeng2022lion} and text\cite{li2022diffusionlm} generation, which shows amazing text-guided image generation capabilities. However, unlike traditional tasks, 3D molecular generation requires us to model and constrain the potential interatomic forces on the target, and also consider the equivariance property. Therefore, we study that the latent space contains equivariant tensors to ensure the equicontrast property, and then model the global and local forces in the latent space.

\textbf{Modeling Interatomic Constraints.} Some methods consider generating molecules in two dimensions, \textit{e.g.}, \cite{dai2018diagnosing, gomez2018automatic, grisoni2020bidirectional} use sequence models such as RNN to generate molecular strings SMILES\cite{weininger1988smiles}, while other models tend to generate molecular graphs composed of atoms, in which chemical bonds are represented by nodes and edges. However, it only uses the fully connected adjacent matrix, thus ignoring the intrinsic topology of the molecular graph. These models all ignore the 3D structural information of molecules, which is particularly critical for the effectiveness and novelty of molecules. Inspired by the success of diffusion models\cite{sohl2015deep} in various tasks\cite{ho2020denoising, song2021denoising, kong2021diffwave}, \cite{hoogeboom2022equivariant} uses diffusion models to generate novel molecular structures in 3D space. However, this only utilizes the fully connected adjacency matrix, ignoring the local constraints between atoms (\textit{i.e.}, chemical bonds (covalent bonds) or van der Waals forces formed between atoms)\cite{huang2022mdm}, and does not integrate the intrinsic topological relationships of the molecular graph and intermolecular forces information, which will greatly reduce the effectiveness and novelty. Our task will consider the above factors to achieve higher expressiveness of the model.
\section{Background}
\subsection{Problem Definition}
\label{ssec:background:define}
We consider generative modeling of spatial molecular geometry in this paper. Define $d$ as the node feature dimension, which contains a dataset of each molecule represented by $\mathcal{G}= \langle \mathrm{\mathbf{x}}, \mathrm{\mathbf{h}} \rangle$, where $\mathrm{x} = (\mathrm{x}_1, \dots, \mathrm{x}_N )\in \mathbb{R}^{N \times 3}$ is the atomic coordinate matrix, $\mathrm{h} = (\mathrm{h}_1, \dots, \mathrm{h}_N )\in \mathbb{R}^{N \times d}$ is the atomic feature matrix, which generally includes atom type and charge. Our task is: \\
\textbf{(I) Unconditional generation.}By parameterizing the generative model $\textit{p}_\theta(\mathcal{G})$ to learn the data distribution of $\mathcal{G}$, we can generate diverse and realistic molecular datasets in space $\hat{\mathcal{G}}$. \\
\textbf{(II) Conditional Control Generation.}Using some molecules $\mathcal{G}$ with desired properties $s$, the parameterized generative model $\textit{p}_\theta(\mathcal{G}|s)$ fits the conditional distribution with the desired properties. The learned model can generate molecules with properties under the condition of given desired properties.

\subsection{Diffussion Process}
Given a molecular geometry $\mathcal{G}_0$, the data is gradually diffused into a predefined noise distribution during the forward diffusion process, and the time is set to \textit{1,\dots,T}. The diffusion model is like more and more particles making increasingly chaotic and irregular motion in space. The process of gradual noise addition in the diffusion model can be formulated as a Markov chain process, which is expressed by the variance table $\beta_1,\dots,\beta_T (\beta_t \in (0,1))$: 
\begin{equation}
\label{eq: diffusion process}
\begin{split}
&q(\mathcal{G}_{1:T}|\mathcal{G}_0)=\prod_{t=1}^{T}q(\mathcal{G}_t|\mathcal{G}_{t-1}),\\
&q(\mathcal{G}_t|\mathcal{G}_{t-1})=\mathcal{N}(\mathcal{G}_t;\sqrt{1-\beta_t}\mathcal{G}_{t-1},\beta_{t}I),
\end{split}
\end{equation}
In the formula, $\mathcal{G}_{t-1}$ is mixed with Gaussian noise to form $\mathcal{G}_t$, where $\beta_t$ is used to control the degree of mixing. In order to simplify the derivation, we set $\bar{\alpha}=\prod_{s=1}^{t}1-\beta_{s}$, so that for any time step $t$, data sampling can be obtained by reparameterization technique with a closed form formula, which is a very important property in the diffusion model: 
\begin{equation}
q\left(\mathcal{G}_t|\mathcal{G}_0\right)=\mathcal{N}\left(\mathcal{G}_t;\sqrt{\bar{\alpha}_{t}}\mathcal{G}_{0},\left(1-\bar{\alpha}_{t}\right) I\right),
\end{equation}
With the increase of the number of steps, $t$ gradually increases, and the final data distribution will be closer to the standard Gaussian distribution, because when $\sqrt{\bar{\alpha_t}} \to 0$, then $(1-\bar{\alpha_t}) \to 1$.
The reverse sampling process of DMs is defined as learning a parameterized reverse denoising process, which aims to gradually denoise the noise variable $\mathcal{G}_{T:1}$ to approximate the initial data distribution $\mathcal{G}_0$ in the target data distribution:
\begin{equation}
\begin{split}
&p_{\theta}\left(\mathcal{G}_{0:{T-1}}|\mathcal{G}_T\right)=\prod_{t-1}^{T}p_{\theta}\left(\mathcal{G}_{t-1}|\mathcal{G}_t\right),\\
&p_{\theta}\left(\mathcal{G}_{t-1}|\mathcal{G}_{t}\right)=\mathcal{N}\left(\mathcal{G}_{t-1};\mu_{\theta}\left(\mathcal{G}_{t},t\right),\sigma^2I\right),
\end{split}
\end{equation}
Where $\mu_{\theta}$ represents the parameterized neural network to approximate the mean of the initial data distribution, $\sigma^2$ represents the user-defined variance, which is generally predefined as 1.
As a latent variable model, the forward process $q(\mathcal{G}_{1:T}|\mathcal{G}_0)$ can be regarded as a fixed posterior, and the backward process $p_{\theta}(\mathcal{G}_{0:T})$ is trained to maximize the variational lower bound of the likelihood of the data $\mathcal{L}_{\mathit{vlb}} = \mathbb{E}_{q(\mathcal{G}_{1:T}|\mathcal{G}_0)}\big[\log\frac{q(\mathcal{G}_T|\mathcal{G}_0)}{p_{\theta}(\mathcal{G}_T)} + \sum_{t=2}^{T}\log\frac{q(\mathcal{G}_{t-1}|\mathcal{G}_0,\mathcal{G}_t)}{p_{\theta}(\mathcal{G}_{t-1}|\mathcal{G}_t)}-\log{p_{\theta}(\mathcal{G}_0|\mathcal{G}_1)}\Big]$. However, we can see that direct optimization of this formula will lead to serious training instability\cite{nichol2021improved}. Instead, \citet{song2019generative,ho2020denoising} gives a simple alternative objective that simplifies to no irrelevant constants:
\begin{equation}
\label{eq:ddpm_loss}
\mathcal{L}_\mathit{DM}=\mathbb{E}_{{\mathcal{G}_0},\mathrm{\epsilon} \sim \mathcal{N}(0,\mathit{\mathbf{I}}),t}\Big[w(t) || \mathrm{\epsilon} - \mathrm{\epsilon}_{\theta}\left(\mathcal{G}_{t}, t\right)||^2 \Big],
\end{equation}
From another perspective, this inverse process predicts the noise part of the data added during the diffusion process at each time step, which is equivalent to the process of moving from low-density areas to high-density areas in the data distribution. The noise-eliminating part is also called \emph{score}\cite{liu2016kernelized}, \textit{e.g.}, the logarithmic density of the data distribution at different time points, which appears in the work of \cite{song2021scorebased}. In order to conveniently reflect $\emph{score}$, we use $\boldsymbol{s}_\theta$ below:
\begin{equation}
\boldsymbol{\mu}_{\theta}\left(\mathcal{G}_{t}, t\right) = \frac{1}{\sqrt{1-{\beta_t}}}\left(\mathcal{G}_{t} + \frac{\beta_{t}}{\sqrt{1-\alpha_{t}}}\boldsymbol{s}_{\theta}\left(\mathcal{G}_{t}, t\right)\right).
\end{equation}
The sampling process of the diffusion model is similar to Langevin dynamics, and $\boldsymbol{s}_{\theta}$ is used as the learning gradient of data density.
\subsection{Equivariance}

The equivariance of Euclidean space is universal for spatial structures, especially molecular structures. Properties of molecular structures, where vector features such as atomic forces or dipole moments should be transformed with respect to the atomic space coordinates \cite{thomas2018tensorfield,fuchs2020se3transformer,batzner2021se}. We can formally express this as follows: the function $\mathcal{F}$ is equivariant with respect to the actions of the group $G$ if $\mathcal{F} \circ S_g(\mathbf{x}) = T_g \circ \mathcal{F}(\mathbf{x}), \forall g \in G$, where $S_g$ and $T_g$ are transformations of the group element g \cite{serre1977linear}. For the special Euclidean group SE(3), i.e. the rotation and translation group in 3D space, its group element $g$ transforms $T_g$ and $S_g$, which can be represented by the translation $t$ and the orthogonal rotation matrix $\mathbf{R}$.

We can intuitively understand that the property characteristics of molecules are invariant in the spatial Euclidean SE(3) group, while the coordinates will be affected by the SE(3) group action transformation, such as the transformation $\mathbf{R}\mathbf{x}+\mathbf{t}=(\mathbf{R}\mathbf{x}_{1} + \mathbf{t},\dots,\mathbf{R}\mathbf{x}_{N}+\mathbf{t})$ after the rotation matrix $R$ and translation $t$. However, we require that this transformation will not affect the properties of the generated molecules, so we must ensure that the learned possibilities are invariant to rotation and translation, which is very important for improving the generalization ability of 3D molecular generative modeling\cite{satorras2021enflow,xu2022geodiff}.
\section{Method}
In this article, we give a detailed description of LMDM\label{method}. As mentioned in the experiment of LDM (Latten Diffusion Model)\cite{rombach2022high}, in the encoder and decoder, the purpose is to reduce the perceptual loss of the original data distribution, and in the diffusion process, it is more about optimizing the semantic loss of the data. Similarly, in the process of molecular generation, in order to further reduce the loss caused by the equivariant properties of molecular geometry, we can first perform equivariant encoding and then perform equivariant decoding; in the intermediate diffusion process, we can model richer "semantic" information (in molecular generation, we think of it as the force between atoms (\textit{e.g.}, covalent bonds or van der Waals forces)), which will make the molecules generated by the model more effective and stable. We use dual equivariant fractional neural networks to model two (long-distance and short-distance) molecular constraints, that is, to reduce the "semantic" information we mentioned earlier, which will help us better maintain the potential properties of the molecule. We provide the overall architecture of the model in \ref{fig:LMDM}.
\subsection{Molecular Autoencoder}
\label{MVAE}
\begin{figure}[!t]
    \centering
    \vspace{3pt}
    \includegraphics[height=0.15\textwidth, width=1\linewidth]{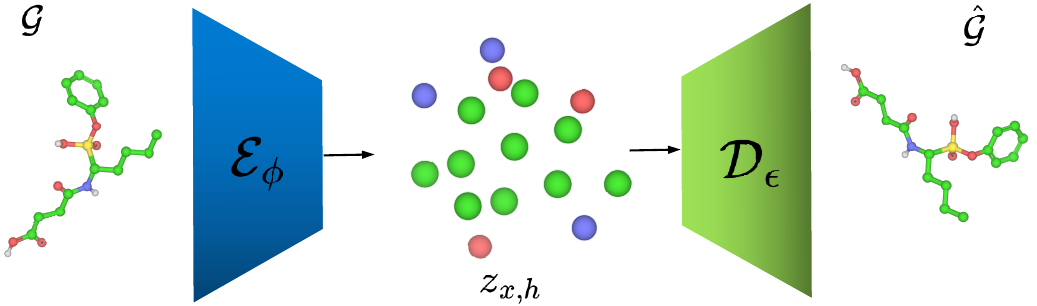} 
    \vspace{-15pt}
    \caption{An overview of the one-stage molecular variational autoencoder. We encode the molecular structure through the encoder $\mathcal{E}_\phi$. Due to the unique properties of EGNN, the encoded latent variables still maintain equivariance on the SE(3) group action. The latent variables are sampled by reparameterization and then decoded by the decoder $\mathcal{D}_\epsilon$ to restore the latent variables to the original molecular structure. As in the first term of \ref{MVAE:lossfunction}, we use $d(\mathcal{G}, \mathcal{\hat{G}})$ to achieve the reconstruction loss, and in order to make the distribution of the latent variable $z_{x,h}$ closer to the prior distribution, making the distribution more regular and smooth, we added the $KL$ regularization term.}
    \label{fig:MVAE}
    \vspace{-10pt}
\end{figure}

We first want to compress the 3D point cloud $\mathcal{G}=\langle \mathrm{x}, \mathrm{h} \rangle \in \mathbb{R}^{(3+d)}$ into a low-dimensional space. In the general autoencoder (AE) framework, the encoder $\mathcal{E}_{\phi}$ maps $\mathcal{G}$ to a low-dimensional latent domain $\mathbf{z}=\mathcal{E}_{\phi}(\mathbf{x},\mathbf{h})$, and the decoder ${\mathcal{D}_{\xi}}$ maps $z$ back to the data domain $\hat{x},\hat{h}=\mathcal{D}_{\xi}(\mathbf{z})$ through training. Figure \ref{fig:MVAE} shows the overall architecture and pipeline of the molecular variational encoder. The entire model minimizes the objective loss function:
\begin{equation}
\label{MVAE:lossfunction}
    \mathcal{L}_{VAE}=\mathbf{\mathit{d}}(\mathcal{D}(\mathcal{E}(\mathcal{G})),\mathcal{G}) + \mathcal{D}_{KL}(q_{\phi}(\mathbf{z}_0 \mid \mathbf{x},\mathbf{h}) {\mid\mid} p{(\mathbf{z}_0)}),
\end{equation}
where $\mathbf{d}$ represents the $p$ norm \textit{e.g.},$ L_p$, and $D_{KL}$ refers to the Kullback-Leibler Divergence. In the formula, $q_{\phi}$ represents the encoded potential distribution, and $p \sim \mathcal{N}(0,I)$ is the standard normal distribution.

In order to maintain the SE(3) group, i.e., rotation and translation equivariance, we usually parameterize the latent space variables into invariant scalar-valued features \cite{kingma2013auto}, which is a considerable difficulty:
\begin{proposition}
\label{prop:se3autoencoding}
    \citep{winter2022unsupervised}Learning autoencoding functions $\mathcal{E}$ and $\mathcal{D}$ to represent geometries $\mathcal{G}$ in scalar-valued (i.e., invariant) latent space \textbf{necessarily} requires an additional \textbf{equivariant} function $\psi$ to store \textbf{suitable} group actions such that $\mathcal{D}(\psi(\mathcal{G}),\mathcal{E}(\mathcal{G})) = T_{\psi(\mathcal{G})} \circ \hat{\mathcal{D}}(\mathcal{E}(\mathcal{G})) = \mathcal{G}$.
\end{proposition}
The method proposed by this proposition is to implement the function $\psi$ in the autoencoder to represent the appropriate group action for encoding, and align the input and output positions for decoding, so as to achieve the purpose of structural reconstruction. In Appendix \ref{app:prop-ae} we give a more detailed explanation and examples. In order to keep the Euclidean group SE(n) action equivariant, \cite{winter2022unsupervised} proposed to set the $\psi$ function to the equivariant orthogonal normal vector of the unit n-dimensional sphere $S^n$.

In our model, we follow the approach used by \cite{xu2023geometric}, constructing latent features into point-structured variables $\mathbf{z}=\langle \mathbf{z}_\mathrm{x}, \mathbf{z}_\mathrm{h} \rangle$, incorporating equivariance into $\mathcal{E}$ and $\mathcal{D}$ instead of applying $\psi$ separately, and jointly representing $\mathbf{z}_\mathrm{x}$ and $\mathbf{z}_\mathrm{h}$, which preserves the $3$-d equivariant and $k$-d invariant latent features $\mathbf{z}_\mathrm{x}$ and $\mathbf{z}_\mathrm{h}$. In the specific implementation, we parameterize $\mathcal{E}$ and $\mathcal{D}$ by using an equivariant graph neural network EGNN\cite{satorras2021enflow}, which has special properties that make the extracted embeddings invariant and equivariant:
\begin{equation}
\label{eq:equivariance:ae}
    \mathbf{R}\mathbf{z}_{\mathrm{x}}+\mathbf{t}, \mathbf{z}_\mathrm{h} = \mathcal{E}_{\phi}(\mathbf{R}\mathbf{x}+\mathbf{t}, \mathbf{h});\mathbf{R}\mathbf{x}+\mathbf{t},\mathbf{h}=\mathcal{D}_\xi(\mathbf{R}\mathbf{z}_\mathrm{x}+\mathbf{t},\mathbf{z}_\mathrm{h}),
\end{equation}
which applies to all orthogonal rotation matrices $\mathbf{R}$ and translations $\textbf{\textit{t}}$. We provide more details on EGNN parameterization in the appendix \ref{app:MVAE detail}. After encoding with $\mathcal{E}_{\phi}$, the latent variables $\mathbf{z}_\mathbf{x}$ are obtained. The latent coordinate variables can play the equivariant effect of the SE(3) group action of the $\psi$ function, maintain the equivariant property of the embedded output, and align the output direction with the input direction. In addition, since the distribution of the latent point variables conforms well to the characteristics of the original distribution, the feature reconstruction of the data can be well achieved.

We are also constraining and optimizing in the objective function of training MVAE. \ref{MVAE:lossfunction} the reconstruction loss is composed of two parts, one is the $L_2$ norm of the atomic coordinate $x$, and the cross entropy of evaluating the atomic type $h$. For the regularization term, we use \textit{ES-reg} \cite{rombach2022high}, adding a little Kullback-Leibler penalty to $q_\phi$ to make it closer to the standard Gaussian distribution; and use \textit{ES-reg}, a regularization of early stopping strategy, to avoid scattered latent space. \textit{ES-reg} also prevents the latent variables from having arbitrarily high variance, so it is more suitable for learning latent distributions\cite{xu2023geometric}.
\subsection{Latent Molecular Diffusion Model}

We then use a dual equivariant fractional neural network \cite{huang2022mdm} as a fractional denoising network to model two levels of edges: local edges within a predefined radius are used to simulate intramolecular forces (\textit{e.g.} covalent bonds) and global edges are used to capture van der Waals forces. And in order to explore the diversity of generation in the network, we add conditional noise, which avoids determining the output of the entire model and improves the diversity of generation. We will describe how the training and sampling phases of LMDM work.
\begin{figure}
    \centering
    \vspace{5pt}
    \includegraphics[height=0.18\textwidth,width=1\linewidth]{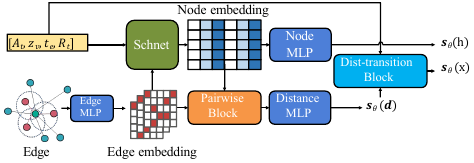}
    \vspace{-15pt}
    \caption{
    Illustration of the specific implementation of the Markov kernel (double equivariant denoising score network). In fact, depending on the local and global edges of the input, we can use it as a local or global equivariant encoder to capture the molecular internal forces in the model and output the expected target score. The implementation of Schnet comes from \cite{schutt2017schnet}.
    }
    \vspace{-15pt}
    \label{fig:kernal}
\end{figure}

As shown in the figure \ref{fig:LMDM}, we first use the \ref{MVAE} mentioned in the previous section to encode the original data into latent space variables, which reduces the dimension of the data while still maintaining the SE(3) group action equivariant property. Then use the dual equivariant fractional neural network \cite{huang2022mdm} as a fractional denoising network to model two levels of edges: local edges within a predefined radius are used to simulate intramolecular forces (\textit{e.g.} covalent bonds) and global edges are used to capture van der Waals forces. And in order to explore the diversity of generation in the network, we add conditional noise, which avoids determining the output of the entire model and improves the diversity of generation. We will describe how the training and sampling stages of LMDM work.

\textbf{The Equivariant Markov Kernels.}\label{Equivariant Markov prop} Since molecular geometry is rotationally and translationally invariant, this property needs to be taken into account when implementing the Markov kernel in the network. The overall architecture is shown in figure\ref{fig:kernal}. In fact, \cite{kohler2020equivariant} proposed an equivariant reversible function that transforms one invariant distribution into another invariant distribution. This theorem also applies to the diffusion model \cite{xu2022geodiff}. If $p(\mathcal{G}_T)$ is invariant and the denoising neural network that learns the parameterization $p(\mathcal{G}_{t-1}|\mathcal{G}_t)$ is equivariant, then the marginal distribution $p(\mathcal{G})$ is also invariant.  We use a double equivariant fractional neural network to implement the equivariant Markov kernel, which satisfies this property. More implementation details will be described in the appendix \ref{app:Markov Kernels}.

\textbf{Edge Construction.} We also need to construct the edges of the atomic nodes in the molecule. In previous work\cite{kohler2020equivariant,hoogeboom2022equivariant}, the fully connected adjacency matrix is input into the equivariant graph neural network. However, this will treat the interatomic effects indiscriminately, but this will ignore the influence of covalent bonds. Therefore, we define the edges within the radius $\tau$ as local edges to simulate covalent bonds, and the remaining edges as global edges to capture the long-range information of van der Waals forces. We generally set the local radius $\tau$ to $2$\AA, because chemical bonds generally do not exceed $2$\AA. In the experiment, we found that if the radius is set too small, this will cause the target distance score to be very close to the diffusion distance score predicted by the score network, but this does not make the generated molecules have good effectiveness and stability, and makes the training of the diffusion process converge more slowly. The atomic features and the coordinates of the local and global edges are input into the dual equivariant network respectively. The local equivariant network simulates the intramolecular forces, such as real chemical bonds, through local edges, while the global equivariant network captures the interaction information between distant atoms, such as van der Waals forces, through global edges.

\textbf{Enhance diversity through variational noise.} We extend the diffusion model to conditional generation by imitating it, and use variational noise to guide the model to learn stronger molecular diversity, that is, adding noise $\epsilon_v$ for conditional generation $p_{\theta}(\mathcal{G}_{0:T-1} \mid \mathcal{G}_T, \epsilon_v)$ to improve diversity. We also use $Schnet$ as the encoder, which outputs the mean $\mu_v$ and the standard deviation $\sigma_v$. We then use the reparameterization technique to obtain the noise $\epsilon_v=\mu_v +\sigma_v^2z,z\in \mathcal{N}(0,I)$. Equation \ref{eq: reverse process}, in the inverse process of the diffusion model, becomes:
\begin{equation}
    \begin{split}
    \label{eq: reverse process}
        &p_{\theta}\left(\mathcal{G}_{0:T-1} \mid \mathcal{G}_{T}, \epsilon_v\right)=\prod_{t-1}^{T}p_{\theta}\left(\mathcal{G}_{t-1} \mid \mathcal{G}_t, \epsilon_v\right),\\
        &p_{\theta}\left(\mathcal{G}_{t-1} \mid \mathcal{G}_t, \epsilon_v\right)=\mathcal{N}\left(\mathcal{G}_{t-1};\mu_{\theta}\left(\mathcal{G}_t,\epsilon_v,t\right),\sigma_t^2\right).
    \end{split}
\end{equation}
\begin{algorithm}[!t]
    \caption{Training Algorithm of LMDM}
    \label{alg:training}
    \begin{algorithmic}[1]
        \STATE{\bf Input:}molecular geometry $\mathcal{G}\langle \mathrm{x},\mathrm{h} \rangle$
        \STATE{\bf Initial:}encoder network $\mathcal{E}_\phi$, decoder network $\mathcal{D}_\xi$, noise encoder $\Phi_v$, global equivariant networks $\Phi_g$, local equivariant networks $\Phi_l$
        \STATE{\bf first Stage: Autoencoder Training}
        \WHILE{$\phi,\xi$ have not converged}
            \STATE $\mu_x, \mu_h, \sigma_x, \sigma_h \leftarrow \mathcal{E}_\phi(\mathrm{x}, \mathrm{h})$ \hfill\COMMENT{Encoding}
            \STATE $\epsilon \in \mathcal{N}(0,I)$
            \STATE Subtract center of gravity from $\epsilon$ in $\epsilon = \left[\epsilon_x, \epsilon_h\right]$
            \STATE $z_x, z_h \leftarrow \epsilon \odot \sigma + \mu$ \hfill\COMMENT{Reparameterization}
            \STATE $\hat{\mathrm{x}}, \hat{\mathrm{h}} \leftarrow \mathcal{D}_\xi(z_x, z_h)$ \hfill\COMMENT{Decoding}
            \STATE $\mathcal{L}_{mvae} = \mathrm{reconstruction}([\hat{\mathrm{x}}, \hat{\mathrm{h}}], \left[\mathrm{x}, \mathrm{h}\right]) + \mathcal{L}_{reg}$
            \STATE $\phi, \xi \leftarrow \operatorname{optimizer}(\mathcal{L}_{mvae};\phi,\xi)$
        \ENDWHILE
        \STATE{\bf Seconde Stage: Diffusion Process Training} 
        \STATE Fix encoder parameters $\phi$
        \REPEAT
        \STATE $z_{x,0}, z_{h,0} \sim q_\phi(z_x, z_h \mid \mathrm{x},\mathrm{h})$\hfill\COMMENT{As lines 5-8}
        \STATE $t \sim \mathcal{U}({1, \dots,T}), \epsilon \in \mathcal{N}(0,I)$
        \STATE Substract center of gravity from $\epsilon_x$ in $\epsilon=[\epsilon_\mathrm{x}, \epsilon_\mathrm{h}]$ 
        \STATE $z_t = \sqrt{\bar{\alpha_t}}z_0 + (1-\bar{\alpha})\epsilon$
        \STATE $\sigma_v, \mu_v = \Phi_v(z_t)$
        \STATE Sample $\eta \in \mathcal{N}(0,I)$, var noise $\eta_v = \mu_v + \sigma_v^2\eta$
        \STATE Regulate:
        \STATE $\mathcal{L}_{vae}=\mathbb{E}_{q_\phi(\eta_v \mid z_t)}(-\mathcal{D}_{KL}(q_{\phi}(\eta_v \mid z_t)) \mid\mid p(\eta))$
        \STATE Prepare gloabl edges $e_g$ and local edges $e_l$
        \STATE $s_\theta(z_t, \eta_v, t) = \Phi_g(z_t, \eta_v, t, e_g) + \Phi_l(z_t, \eta_v, t, e_l)$
        \STATE Take gradient descent step on
        \STATE $\nabla_\theta\mid\mid s_\theta(z_t, \eta_v,t) - \nabla_{z_t}\mathrm{log}q_\Phi(z_t \mid z_0) \mid\mid^2 + \mathcal{L}_{vae}$
        \UNTIL $\Phi_v, \Phi_g, \Phi_l$ have Converged
    \end{algorithmic}
\end{algorithm}

In the inverse process of the diffusion model, we apply an special sampling strategy. In the experiment, when sampling $z_v$ from the uniform distribution $\mathcal{U}(-1,+1)$, the generation effect of the model is significantly improved.
\begin{algorithm}
\caption{Sampling Algorithm of LMDM}
\label{alg:sampling}
\begin{algorithmic}[1]
     \STATE {\bf input:} decoder network $\mathcal{D}_\xi$, learned global and local equivariant network $\Phi_g$, $\Phi_l$
    \STATE Sample $z_T \sim \mathcal{N}(0,I)$
    \FOR{$t$ in $T,\,T-1,\cdots, 1$}
        \STATE Sample $\epsilon \sim \mathcal{N}(0,I)$ if $t>1$, else $\epsilon=0$
        \STATE Shift $\mathrm{x}_t$ to zero COM in $z_t=[x_t,h_t]$
        \STATE Prepare gloabl edges $e_g$ and local edges $e_l$
        \STATE Sample $\eta_v \sim \mathcal{N}(0,I)$
        \STATE $s_\theta(z_t, \eta_v, t)=\Phi_g(z_t, \eta_v, t,e_g)+\Phi_l(z_t,\eta_v,t,e_l)$
        \STATE $\mu_\theta(z_t,\eta_v,t)=\frac{1}{\sqrt{1-\beta_t}}\left(z_t+\frac{\beta_t}{\sqrt{1-\bar{\alpha}_t}}s_\theta(z_t,\eta_v,t)\right)$
        \STATE $z_{t-1} = \mu_\theta(z_t,\eta_v,t)+\sigma_t\epsilon$
    \ENDFOR
    \STATE $\mathbf{x, h} \sim p_\xi(x,h \mid z_{x,0}, z_{h,0})$\hfill\COMMENT{Decoding}
    \STATE \textbf{return} $\mathcal{G}_0$ to obtain $\mathbf{x,h}$
\end{algorithmic}
\end{algorithm}

\subsection{Taining And Sampling}
In the experiment, we found that the training method of previous work \cite{xu2023geometric}, in our LMDM, makes the model difficult to generalize on the validation set and the convergence speed is reduced. Therefore, we still use a two-stage approach, first training the molecular autoencoder, then fixing the encoder and training the diffusion process.

During the sampling process, the atomic coordinates $\mathrm{x}_t$ may violate the equal variance requirement \cite{huang2022mdm}, so we need to better estimate the gradient target of the potential distribution of molecular coordinates. We sample $\mathrm{x}_t$ at the pairwise distance $\mathrm{d}_{ij}$:
\begin{equation}
    \nabla_{\mathrm{\tilde{x}}}\log q_{\Phi}(\mathrm{\tilde{x}}_i|\mathrm{x}_i)=\sum_{j \in N(i)}\frac{\nabla_{\tilde{d}_{ij}}q_{\Phi}(\tilde{\mathrm{d}}_{ij} \mid \mathrm{d}_{ij}) \cdot (\mathrm{x}_i - \mathrm{x}_j)}{\mathrm{d}_{ij}},
\end{equation}
where $\tilde{\mathrm{x}}$ refers to the coordinate of the diffusing atom at $z_t$ and $\tilde{\mathrm{d}}$ denotes the corresponding diffusion distance. We make an approximate estimate of the diffusion distance gradient of the potential distribution as:
\begin{equation}
    \nabla_{\mathrm{\tilde{d}}}\log q_\Phi(\mathrm{\tilde{d}} \mid \mathrm{d}) \sim \frac{-\sqrt{\hat{\alpha}_t}(\tilde{\mathrm{d}}-\mathrm{d})}{1-\bar{\alpha}_t}.
\end{equation}

Just like the diffusion model \cite{ho2020denoising}, our training objective is obtained by optimizing the variational lower bound of the negative log-likelihood (ELBO). We obtain our objective function in the same way and simplify the training objective: 
\begin{equation} 
    \mathcal{L}_{t}=\mathbb{E}_{\mathcal{G}_0}[\gamma\mid\mid s_\theta(z_t,\eta_v,t) - \nabla_{z_t}\log (z_t \mid z_0)\mid\mid^2], 
\end{equation}
Where $\gamma =\frac{\beta_t^2}{2(1-\beta_t)(1-\bar{\alpha}_t)\sigma_t^2}$ represents the weight and $\sigma_t^2$ denotes the user-defined variance. However, based on experience, we ignore $\gamma$ and the simplified objective performs better. we provide more information on the derivation of the training objective in the Appendix\ref{app:prop-diffusion}.

\begin{figure*}[!t]
    \centering
    \includegraphics[width=1\linewidth, height=0.15\textwidth]{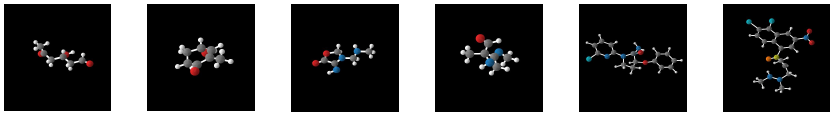}
    \caption{Molecules generated by LMDM trained on QM9 (left four) and DRUG (right two).}
    \label{fig:uncondition}
\end{figure*}
Combined with our previous variational noise $\mathbf{KL}$ loss, we get the final training objective:
\begin{equation}
    \mathcal{L}=\sum_{t=2}^T(\mathcal{L}_t+\mathcal{L}_{vae}),
\end{equation}
This simplified objective is equivalent to learning $s_\theta$ by sampling the diffusing molecules $z_t$ with a time step of $t$ and using the logarithmic density gradient of the data distribution.

Algorithm \ref{alg:training} shows the complete training process. We first train the AE to add regularization, and then train the latent diffusion process using the latent code encoded by the pre-trained encoder. Each numerator of the fused random time step $t \sim \mathcal{U}(1, T)$ will be perturbed by the noise $\epsilon$. To ensure the invariance of $\epsilon$, we use the zero center of mass (COM) method of \cite{kohler2020equivariant} to ensure that $p(\mathcal{G}_T)$ remains unchanged. And extend $p(\mathcal{G}_T)$ to an isotropic Gaussian, $\epsilon$ is invariant to rotation and translation around the zero COM.

The potential code distribution obtained by the diffusion process is defined as the residual distribution \cite{xu2023geometric}: $p_{\theta,\xi}(\mathrm{x,h,z_x,z_h})=p_\theta(\mathrm{z_x,z_h})p_\xi(\mathrm{x,h \mid z_x, z_h})$. where $p_\theta$ represents the diffusion model that models the potential code distribution, and $p_\xi$ represents the decoder. We can first sample equivariant potential codes from $p_\theta$, and then use $p_\xi$ to convert them to the molecular structure of the original space. The algorithm \ref{alg:sampling} provides pseudo code for the sampling process.
\section{Experiments}
In this section, we report the training results of the LMDM model on two benchmark datasets (QM9 \cite{ramakrishnan2014quantum} and GEOM \cite{axelrod2022geom}), which show that the proposed LMDM significantly outperforms multiple state-of-the-art - the-art (SOTA) 3D molecule generation method. We conducted unconditional generation experiments on two data sets, as well as conditional control generation experiments, to evaluate the ability of LMDM to generate molecules with desired properties. I also conducted ablation experiments in the appendix\ref{app:exp-ablation} to compare the performance of the encoder in modeling covalent bonds or van der Waals forces between molecules in the latent diffusion process with or without KL regularization.

\subsection{Molecular Geometry Generation.}
\textbf{Dataset} We use QM9\cite{ramakrishnan2014quantum} and GEOMDrug\cite{axelrod2022geom} to evaluate the performance of LMDM. QM9 contains more than 130k molecules, each containing an average of 18 atoms. GEOM-Drug contains 290K molecules, each containing an average of 46 atoms. We will introduce more details about these two datasets and the division settings of the training set and the validation set in the appendix\ref{app:dataset}.

\textbf{Baselines and setup.} We compare LMDM with two generative models, including EDM\cite{hoogeboom2022equivariant} and GeoLDM\cite{xu2023geometric}, and an autoregressive model G-Schnet\cite{gebauer2019symmetry}. For these three models, we use the published pre-trained models for evaluation and comparison. Due to the data processing script and corresponding configuration file of G-Schnet on the GEOM-Drug dataset, it is impossible to reproduce G-Schnet on GEOM-Drug. We also studied the impact of whether the potential code distribution is aligned with the standard normal distribution on model performance, and removed $\mathcal{D}_{KL}$ in the MVAE stage to study its impact.

\textbf{Evaluation Metrics.} Following previous work on 3D molecule generation\cite{gebauer2019symmetry,satorras2021enflow,hoogeboom2022equivariant,wu2022diffusionbased}, We measure the generation performance of our model using four metrics:
\begin{itemize}\small
    \item \textbf{Validity:} The percentage of molecules generated by the model that follow the chemical valence rules specified by RDkit;
    \item \textbf{Uniqueness:} the percentage of unique and valid molecules among the molecules generated by the model;
    \item \textbf{Novelty:} the percentage of unique molecules generated that are not in the training set;
    \item \textbf{Stability:} the percentage of molecules generated without ions in the total number of all generated molecules.
\end{itemize}
\begin{figure*}[!t]
    \centering
    \includegraphics[width=0.85\linewidth, height=0.18\textwidth]{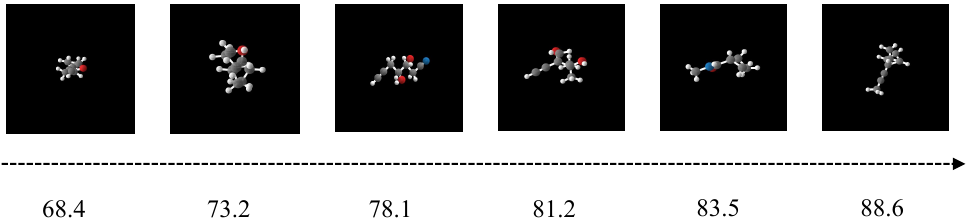}
    \caption{Molecules generated by conditional LMDM. We conduct controllable generation with interpolation among different Polariz-ability $\alpha$ values with the same reparametrization noise $\epsilon$. The given $\alpha$ values are provided at the bottom.}
    \label{fig:condition_alpha}
    \vspace{-7pt}
\end{figure*}
\begin{table*}[htb]\small
\centering
\caption{The comparison over 10k generated molecules of LMDM and baseline models on molecular geometry generation task. $\uparrow$ means that higher the values, better the performance of the model.  \label{Tab:1}}
\scalebox{0.8}{
\begin{tabular}{lcccccccc}\\\toprule 
& \multicolumn{4}{c}{QM9} & \multicolumn{4}{c}{GEOM}\\
~ &  Validity $(\%)$ $\uparrow$ &  Uniqueness $(\%)$ $\uparrow$ &  Novelty $(\%)$ $\uparrow$ &  Stability $(\%)$ $\uparrow$ &  Validity $(\%)$ $\uparrow$ &  Uniqueness $(\%)$ $\uparrow$ &  Novelty $(\%)$ $\uparrow$ &  Stability $(\%)$ $\uparrow$\\\midrule
G-Schnet & 85.9 & 80.9 & 57.6 & 85.6 & - & - & - & -\\
EDM & 91.7 & 90.5 & 59.9 & 91.1 & 68.6 & 68.6 & 68.6 & 13.7\\
GeoLDM & 93.4 & \textbf{98.7} & - & 88.9 & 99.3 & - & - & -\\
\midrule
LMDM-KL & 96.2 & 92.1 & 89.2 & 87.5 & 99.2 & 99.1 & \textbf{99.2} & 52.4\\
LMDM  & \textbf{98.8} & 95.2 & \textbf{92.1} & \textbf{90.8} & \textbf{99.5} & \textbf{99.2} & 99.1 & \textbf{63.4} \\
\bottomrule
\end{tabular}}
\vspace{-7pt}
\end{table*}

\textbf{Results and Analysis} From Table\ref{Tab:1}, we can see that LMDM outperforms all baseline models by generating 10k molecular samples from the above models to calculate the evaluation indicators. Note that on the GEOM-Drug dataset, the atomic-level stability of the dataset itself is as high as 86.5\%, but the molecular-level stability is close to 0\%. This is because GEOM-Drug is a drug molecule, which usually contains larger and more complex physical structures, and will accumulate larger errors in predictions based on interatomic distances and atom pair types. our proposed method demonstrates its advantage in generating high-quality molecules, which is more evident when the generated molecules contain a large number of atoms. The reason behind the advantages of the model is largely due to the fact that we perform equivariant latent encoding on the original molecular data, keep the distribution of the original data as much as possible and make the latent variable distribution regular and smooth, and use a dual equivariant fractional neural network to capture the interatomic forces (\textit{e.g.}, chemical bonds and van der Waals forces) during the diffusion process. With these advantages, the latent diffusion process successfully captures the structural patterns between atoms at different distances, so that the model can make progress in the stability modeling of large molecules such as GEOM-Drug. We also noticed that the model trained with the KL term did not perform as well as the LMDM in terms of results. This is because the restriction of the KL term may cause the output latent variable distribution to deviate from the data prior distribution, which will lead to distribution deviations when constructing intermolecular forces. For example, the distance distribution of the atomic pair 2\text{\AA} may be forcibly controlled to a distance distribution that conforms to the standard normal distribution.

\subsection{Conditional Molecular Generation}
\textbf{Baseline and Setup}In this section, we control various properties about the molecule as conditions to generate molecules with our target properties. We train a model with six properties on the QM9 dataset: polarizability $\mathbf{\alpha}$, \textbf{HOMO} $\epsilon_{\mathrm{HOMO}}$, \textbf{LUMO} $\epsilon_{\mathrm{LUMO}}$, \textbf{HOMO-LUMO} gap $\epsilon_{\mathrm{gap}}$, Dipole moment $\mu$ and $C_v$. We achieve conditional probability generation by connecting these conditions $\mathrm{c}$ with atomic features to obtain $p(\mathcal{G}_{t-1}|\mathcal{G}_t,\mathrm{c})$.
\begin{table}[htb]\small
\vspace{-7pt}
    \centering
    \caption{The results of conditional molecular generation on QM9
dataset. $\downarrow$ means the lower the values, the better the model incor-
porates the targeted properties.}
\scalebox{0.8}{
    {\begin{tabular}{lcccccc} \\\toprule
    Methods & $\alpha$ $\downarrow$ & $\epsilon_{\text{{gap}}}$ $\downarrow$ & $\epsilon_{\text{{HOMO}}}$ $\downarrow$ & $\epsilon_{\text{{LUMO}}}$ $\downarrow$ & $\mu$ $\downarrow$& $C_v$ $\downarrow$ \\\midrule
    Naive (U-bound) & 9.013 & 1.472 & 0.645 & 1.457 & 1.616 & 6.857\\
    \#Atoms & 3.862 & 0.866 & 0.426 & 0.813 & 1.053 & 1.971\\
    \midrule
    GeoLDM & 2.370 & 0.587 & 0.340 & 0.522 & 1.108 & 1.025\\
    \rowcolor{lightgray}
    LMDM & 1.621 & 0.068 & 0.041 & 0.047 & 1.249 & 1.726\\
    \midrule
    QM9(L-bound) & 0.100 & 0.064 & 0.039 & 0.036 & 0.043 & 0.040\\\bottomrule
    \end{tabular}}}
    \label{Tab:2}
\end{table}

\textbf{Evaluation Metrics} Following previous work\cite{hoogeboom2022equivariantdiffusionmoleculegeneration}, we train a property classifier (PC)\cite{satorras2021enflow}. The QM9 dataset is divided into two parts, each containing 50k samples. The first half $D_1$ is used to train the property classifier, and the other half $D_2$ is used to train the generative model. The classifier $\phi_c$ then evaluates the conditional generated samples by the mean absolute error (MAE) between the predicted attribute values and the true attribute values.

\textbf{Result and Analysis}
The conditional generation results of LMDM are given in Table \ref{Tab:2}. We can notice that, except for the indicators $\mu$ and $C_v$, almost all attributes can exceed "Naive", "\#Atoms" and GeoLDM. This result shows that LMDM can well integrate the conditional attribute information into the generated samples, our model can well fit the distribution of $D_2$, and can generate molecules with target attributes. We also interpolate the molecules generated by conditional LMDM when different polarizabilities $\alpha$ in Figure \ref{fig:condition_alpha}, which is consistent with the expectation that larger polarizabilities $\alpha$ have smaller isotropic shapes.
\section{Conclusion}
In this study, we proposed a new latent diffusion model LMDM. Instead of operating on high-dimensional, multimodal atomic features, we learn the diffusion process through continuous, latent space to overcome the limitations of excessive time and space complexity in the original space. By constructing a point structure with invariant and equivariant tensors to form molecular latent variables that preserve rotation and translation, and modeling molecular topology at long and short distances, covalent bonds or van der Waals forces between molecules are captured. Our experimental results show that it has significantly better capabilities in simulating chemically real molecules and can generate molecules with desired properties. In the future, our model can be applied to more complex scenarios, such as protein and drug discovery.

\bibliography{main}

\begin{thebibliography}{53}
\providecommand{\natexlab}[1]{#1}
\providecommand{\url}[1]{\texttt{#1}}
\expandafter\ifx\csname urlstyle\endcsname\relax
  \providecommand{\doi}[1]{doi: #1}\else
  \providecommand{\doi}{doi: \begingroup \urlstyle{rm}\Url}\fi

\bibitem[Anderson et~al.(2019)Anderson, Hy, and Kondor]{anderson2019cormorant}
Anderson, B., Hy, T.~S., and Kondor, R.
\newblock Cormorant: Covariant molecular neural networks.
\newblock \emph{Advances in neural information processing systems}, 32, 2019.

\bibitem[Aneja et~al.(2021)Aneja, Schwing, Kautz, and Vahdat]{aneja2021contrastive}
Aneja, J., Schwing, A., Kautz, J., and Vahdat, A.
\newblock A contrastive learning approach for training variational autoencoder priors.
\newblock \emph{Advances in neural information processing systems}, 34:\penalty0 480--493, 2021.

\bibitem[Axelrod \& Gomez-Bombarelli(2022)Axelrod and Gomez-Bombarelli]{axelrod2022geom}
Axelrod, S. and Gomez-Bombarelli, R.
\newblock Geom, energy-annotated molecular conformations for property prediction and molecular generation.
\newblock \emph{Scientific Data}, 9\penalty0 (1):\penalty0 1--14, 2022.

\bibitem[Batzner et~al.(2021)Batzner, Smidt, Sun, Mailoa, Kornbluth, Molinari, and Kozinsky]{batzner2021se}
Batzner, S., Smidt, T.~E., Sun, L., Mailoa, J.~P., Kornbluth, M., Molinari, N., and Kozinsky, B.
\newblock Se (3)-equivariant graph neural networks for data-efficient and accurate interatomic potentials.
\newblock \emph{arXiv preprint arXiv:2101.03164}, 2021.

\bibitem[Dai \& Wipf(2019)Dai and Wipf]{dai2018diagnosing}
Dai, B. and Wipf, D.
\newblock Diagnosing and enhancing {VAE} models.
\newblock In \emph{International Conference on Learning Representations}, 2019.
\newblock URL \url{https://openreview.net/forum?id=B1e0X3C9tQ}.

\bibitem[Fuchs et~al.(2020)Fuchs, Worrall, Fischer, and Welling]{fuchs2020se3transformer}
Fuchs, F., Worrall, D.~E., Fischer, V., and Welling, M.
\newblock Se(3)-transformers: 3d roto-translation equivariant attention networks.
\newblock In \emph{Advances in Neural Information Processing Systems 33: Annual Conference on Neural Information Processing Systems 2020, {NeurIPS}}, 2020.

\bibitem[Gebauer et~al.(2019)Gebauer, Gastegger, and Sch{\"u}tt]{gebauer2019symmetry}
Gebauer, N., Gastegger, M., and Sch{\"u}tt, K.
\newblock Symmetry-adapted generation of 3d point sets for the targeted discovery of molecules.
\newblock \emph{Advances in neural information processing systems}, 32, 2019.

\bibitem[Gebauer et~al.(2020)Gebauer, Gastegger, and Schütt]{gebauer2020symmetryadaptedgeneration3dpoint}
Gebauer, N. W.~A., Gastegger, M., and Schütt, K.~T.
\newblock Symmetry-adapted generation of 3d point sets for the targeted discovery of molecules, 2020.
\newblock URL \url{https://arxiv.org/abs/1906.00957}.

\bibitem[G{\'o}mez-Bombarelli et~al.(2018)G{\'o}mez-Bombarelli, Wei, Duvenaud, Hern{\'a}ndez-Lobato, S{\'a}nchez-Lengeling, Sheberla, Aguilera-Iparraguirre, Hirzel, Adams, and Aspuru-Guzik]{gomez2018automatic}
G{\'o}mez-Bombarelli, R., Wei, J.~N., Duvenaud, D., Hern{\'a}ndez-Lobato, J.~M., S{\'a}nchez-Lengeling, B., Sheberla, D., Aguilera-Iparraguirre, J., Hirzel, T.~D., Adams, R.~P., and Aspuru-Guzik, A.
\newblock Automatic chemical design using a data-driven continuous representation of molecules.
\newblock \emph{ACS central science}, 4\penalty0 (2):\penalty0 268--276, 2018.

\bibitem[Grisoni et~al.(2020)Grisoni, Moret, Lingwood, and Schneider]{grisoni2020bidirectional}
Grisoni, F., Moret, M., Lingwood, R., and Schneider, G.
\newblock Bidirectional molecule generation with recurrent neural networks.
\newblock \emph{Journal of chemical information and modeling}, 60\penalty0 (3):\penalty0 1175--1183, 2020.

\bibitem[Ho et~al.(2020)Ho, Jain, and Abbeel]{ho2020denoising}
Ho, J., Jain, A., and Abbeel, P.
\newblock Denoising diffusion probabilistic models.
\newblock \emph{arXiv preprint arXiv:2006.11239}, 2020.

\bibitem[Hoogeboom et~al.(2022{\natexlab{a}})Hoogeboom, Satorras, Vignac, and Welling]{hoogeboom2022equivariant}
Hoogeboom, E., Satorras, V.~G., Vignac, C., and Welling, M.
\newblock Equivariant diffusion for molecule generation in 3d.
\newblock In \emph{International Conference on Machine Learning}, pp.\  8867--8887. PMLR, 2022{\natexlab{a}}.

\bibitem[Hoogeboom et~al.(2022{\natexlab{b}})Hoogeboom, Satorras, Vignac, and Welling]{hoogeboom2022equivariantdiffusionmoleculegeneration}
Hoogeboom, E., Satorras, V.~G., Vignac, C., and Welling, M.
\newblock Equivariant diffusion for molecule generation in 3d, 2022{\natexlab{b}}.
\newblock URL \url{https://arxiv.org/abs/2203.17003}.

\bibitem[Huang et~al.(2022)Huang, Zhang, Xu, and Wong]{huang2022mdm}
Huang, L., Zhang, H., Xu, T., and Wong, K.-C.
\newblock Mdm: Molecular diffusion model for 3d molecule generation.
\newblock \emph{arXiv preprint arXiv:2209.05710}, 2022.

\bibitem[Jing et~al.(2021)Jing, Eismann, Suriana, Townshend, and Dror]{jing2021learningproteinstructuregeometric}
Jing, B., Eismann, S., Suriana, P., Townshend, R. J.~L., and Dror, R.
\newblock Learning from protein structure with geometric vector perceptrons, 2021.
\newblock URL \url{https://arxiv.org/abs/2009.01411}.

\bibitem[Jo et~al.(2022)Jo, Lee, and Hwang]{jo2022score}
Jo, J., Lee, S., and Hwang, S.~J.
\newblock Score-based generative modeling of graphs via the system of stochastic differential equations.
\newblock \emph{arXiv preprint arXiv:2202.02514}, 2022.

\bibitem[Jumper et~al.(2021)Jumper, Evans, Pritzel, Green, Figurnov, Ronneberger, Tunyasuvunakool, Bates, {\v{Z}}{\'\i}dek, Potapenko, et~al.]{jumper2021highly}
Jumper, J., Evans, R., Pritzel, A., Green, T., Figurnov, M., Ronneberger, O., Tunyasuvunakool, K., Bates, R., {\v{Z}}{\'\i}dek, A., Potapenko, A., et~al.
\newblock Highly accurate protein structure prediction with alphafold.
\newblock \emph{Nature}, 596\penalty0 (7873):\penalty0 583--589, 2021.

\bibitem[Kingma \& Ba(2015)Kingma and Ba]{DBLP:journals/corr/KingmaB14}
Kingma, D.~P. and Ba, J.
\newblock Adam: {A} method for stochastic optimization.
\newblock In Bengio, Y. and LeCun, Y. (eds.), \emph{3rd International Conference on Learning Representations, {ICLR} 2015, San Diego, CA, USA, May 7-9, 2015, Conference Track Proceedings}, 2015.
\newblock URL \url{http://arxiv.org/abs/1412.6980}.

\bibitem[Kingma \& Welling(2013)Kingma and Welling]{kingma2013auto}
Kingma, D.~P. and Welling, M.
\newblock Auto-encoding variational bayes.
\newblock In \emph{2nd International Conference on Learning Representations}, 2013.

\bibitem[Klicpera et~al.(2020)Klicpera, Gro{\ss}, and G{\"{u}}nnemann]{klicpera2020dimenet}
Klicpera, J., Gro{\ss}, J., and G{\"{u}}nnemann, S.
\newblock Directional message passing for molecular graphs.
\newblock In \emph{8th International Conference on Learning Representations, {ICLR}}, 2020.

\bibitem[K{\"o}hler et~al.(2020)K{\"o}hler, Klein, and No{\'e}]{kohler2020equivariant}
K{\"o}hler, J., Klein, L., and No{\'e}, F.
\newblock Equivariant flows: exact likelihood generative learning for symmetric densities.
\newblock In \emph{International conference on machine learning}, pp.\  5361--5370. PMLR, 2020.

\bibitem[Kong et~al.(2021)Kong, Ping, Huang, Zhao, and Catanzaro]{kong2021diffwave}
Kong, Z., Ping, W., Huang, J., Zhao, K., and Catanzaro, B.
\newblock Diffwave: A versatile diffusion model for audio synthesis.
\newblock In \emph{International Conference on Learning Representations}, 2021.
\newblock URL \url{https://openreview.net/forum?id=a-xFK8Ymz5J}.

\bibitem[Li et~al.(2022)Li, Thickstun, Gulrajani, Liang, and Hashimoto]{li2022diffusionlm}
Li, X.~L., Thickstun, J., Gulrajani, I., Liang, P., and Hashimoto, T.
\newblock Diffusion-{LM} improves controllable text generation.
\newblock In Oh, A.~H., Agarwal, A., Belgrave, D., and Cho, K. (eds.), \emph{Advances in Neural Information Processing Systems}, 2022.
\newblock URL \url{https://openreview.net/forum?id=3s9IrEsjLyk}.

\bibitem[Lin et~al.(2022)Lin, Huang, Liu, Li, Ji, and Li]{lin2022diffbp}
Lin, H., Huang, Y., Liu, M., Li, X., Ji, S., and Li, S.~Z.
\newblock Diffbp: Generative diffusion of 3d molecules for target protein binding.
\newblock \emph{arXiv preprint arXiv:2211.11214}, 2022.

\bibitem[Liu et~al.(2016)Liu, Lee, and Jordan]{liu2016kernelized}
Liu, Q., Lee, J., and Jordan, M.
\newblock A kernelized stein discrepancy for goodness-of-fit tests.
\newblock In \emph{International conference on machine learning}, pp.\  276--284. PMLR, 2016.

\bibitem[Luo et~al.(2022)Luo, Su, Peng, Wang, Peng, and Ma]{luo2022antigenspecific}
Luo, S., Su, Y., Peng, X., Wang, S., Peng, J., and Ma, J.
\newblock Antigen-specific antibody design and optimization with diffusion-based generative models for protein structures.
\newblock In Oh, A.~H., Agarwal, A., Belgrave, D., and Cho, K. (eds.), \emph{Advances in Neural Information Processing Systems}, 2022.
\newblock URL \url{https://openreview.net/forum?id=jSorGn2Tjg}.

\bibitem[Ma et~al.(2019)Ma, Zhou, Li, Neubig, and Hovy]{ma2019flowseq}
Ma, X., Zhou, C., Li, X., Neubig, G., and Hovy, E.
\newblock {F}low{S}eq: Non-autoregressive conditional sequence generation with generative flow.
\newblock In \emph{Proceedings of the 2019 Conference on Empirical Methods in Natural Language Processing and the 9th International Joint Conference on Natural Language Processing (EMNLP-IJCNLP)}, pp.\  4282--4292, Hong Kong, China, November 2019. Association for Computational Linguistics.
\newblock \doi{10.18653/v1/D19-1437}.
\newblock URL \url{https://aclanthology.org/D19-1437}.

\bibitem[Nichol \& Dhariwal(2021)Nichol and Dhariwal]{nichol2021improved}
Nichol, A.~Q. and Dhariwal, P.
\newblock Improved denoising diffusion probabilistic models.
\newblock In \emph{International Conference on Machine Learning}, pp.\  8162--8171. PMLR, 2021.

\bibitem[Pereira et~al.(2016)Pereira, Caffarena, and Dos~Santos]{pereira2016boosting}
Pereira, J.~C., Caffarena, E.~R., and Dos~Santos, C.~N.
\newblock Boosting docking-based virtual screening with deep learning.
\newblock \emph{Journal of chemical information and modeling}, 56\penalty0 (12):\penalty0 2495--2506, 2016.

\bibitem[Ramakrishnan et~al.(2014)Ramakrishnan, Dral, Rupp, and Von~Lilienfeld]{ramakrishnan2014quantum}
Ramakrishnan, R., Dral, P.~O., Rupp, M., and Von~Lilienfeld, O.~A.
\newblock Quantum chemistry structures and properties of 134 kilo molecules.
\newblock \emph{Scientific data}, 1\penalty0 (1):\penalty0 1--7, 2014.

\bibitem[Razavi et~al.(2019)Razavi, Van~den Oord, and Vinyals]{razavi2019generating}
Razavi, A., Van~den Oord, A., and Vinyals, O.
\newblock Generating diverse high-fidelity images with vq-vae-2.
\newblock \emph{Advances in neural information processing systems}, 32, 2019.

\bibitem[Rombach et~al.(2022)Rombach, Blattmann, Lorenz, Esser, and Ommer]{rombach2022high}
Rombach, R., Blattmann, A., Lorenz, D., Esser, P., and Ommer, B.
\newblock High-resolution image synthesis with latent diffusion models.
\newblock In \emph{Proceedings of the IEEE/CVF Conference on Computer Vision and Pattern Recognition}, pp.\  10684--10695, 2022.

\bibitem[Satorras et~al.(2021)Satorras, Hoogeboom, Fuchs, Posner, and Welling]{satorras2021enflow}
Satorras, V.~G., Hoogeboom, E., Fuchs, F.~B., Posner, I., and Welling, M.
\newblock E (n) equivariant normalizing flows for molecule generation in 3d.
\newblock \emph{arXiv preprint arXiv:2105.09016}, 2021.

\bibitem[Satorras et~al.(2022)Satorras, Hoogeboom, and Welling]{satorras2022enequivariantgraphneural}
Satorras, V.~G., Hoogeboom, E., and Welling, M.
\newblock E(n) equivariant graph neural networks, 2022.
\newblock URL \url{https://arxiv.org/abs/2102.09844}.

\bibitem[Sch\"{u}tt et~al.(2017)Sch\"{u}tt, Kindermans, Sauceda~Felix, Chmiela, Tkatchenko, and M\"{u}ller]{schutt2017schnet}
Sch\"{u}tt, K., Kindermans, P.-J., Sauceda~Felix, H.~E., Chmiela, S., Tkatchenko, A., and M\"{u}ller, K.-R.
\newblock Schnet: A continuous-filter convolutional neural network for modeling quantum interactions.
\newblock In \emph{Advances in Neural Information Processing Systems}, pp.\  991--1001. Curran Associates, Inc., 2017.

\bibitem[Serre et~al.(1977)]{serre1977linear}
Serre, J.-P. et~al.
\newblock \emph{Linear representations of finite groups}, volume~42.
\newblock Springer, 1977.

\bibitem[Simonovsky \& Komodakis(2018)Simonovsky and Komodakis]{simonovsky2018graphvae}
Simonovsky, M. and Komodakis, N.
\newblock Graphvae: Towards generation of small graphs using variational autoencoders.
\newblock In \emph{International conference on artificial neural networks}, pp.\  412--422. Springer, 2018.

\bibitem[Sohl-Dickstein et~al.(2015)Sohl-Dickstein, Weiss, Maheswaranathan, and Ganguli]{sohl2015deep}
Sohl-Dickstein, J., Weiss, E., Maheswaranathan, N., and Ganguli, S.
\newblock Deep unsupervised learning using nonequilibrium thermodynamics.
\newblock In \emph{International Conference on Machine Learning}, pp.\  2256--2265. PMLR, 2015.

\bibitem[Song et~al.(2021{\natexlab{a}})Song, Meng, and Ermon]{song2021denoising}
Song, J., Meng, C., and Ermon, S.
\newblock Denoising diffusion implicit models.
\newblock In \emph{International Conference on Learning Representations}, 2021{\natexlab{a}}.
\newblock URL \url{https://openreview.net/forum?id=St1giarCHLP}.

\bibitem[Song \& Ermon(2019)Song and Ermon]{song2019generative}
Song, Y. and Ermon, S.
\newblock Generative modeling by estimating gradients of the data distribution.
\newblock In \emph{Advances in Neural Information Processing Systems}, pp.\  11918--11930, 2019.

\bibitem[Song et~al.(2021{\natexlab{b}})Song, Sohl-Dickstein, Kingma, Kumar, Ermon, and Poole]{song2021scorebased}
Song, Y., Sohl-Dickstein, J., Kingma, D.~P., Kumar, A., Ermon, S., and Poole, B.
\newblock Score-based generative modeling through stochastic differential equations.
\newblock In \emph{International Conference on Learning Representations}, 2021{\natexlab{b}}.

\bibitem[Thomas et~al.(2018)Thomas, Smidt, Kearnes, Yang, Li, Kohlhoff, and Riley]{thomas2018tensorfield}
Thomas, N., Smidt, T., Kearnes, S.~M., Yang, L., Li, L., Kohlhoff, K., and Riley, P.
\newblock Tensor field networks: Rotation- and translation-equivariant neural networks for 3d point clouds.
\newblock \emph{CoRR}, abs/1802.08219, 2018.

\bibitem[Townshend et~al.(2021)Townshend, V{\"o}gele, Suriana, Derry, Powers, Laloudakis, Balachandar, Jing, Anderson, Eismann, Kondor, Altman, and Dror]{townshend2021atomd}
Townshend, R. J.~L., V{\"o}gele, M., Suriana, P.~A., Derry, A., Powers, A., Laloudakis, Y., Balachandar, S., Jing, B., Anderson, B.~M., Eismann, S., Kondor, R., Altman, R., and Dror, R.~O.
\newblock {ATOM}3d: Tasks on molecules in three dimensions.
\newblock In \emph{Thirty-fifth Conference on Neural Information Processing Systems Datasets and Benchmarks Track (Round 1)}, 2021.
\newblock URL \url{https://openreview.net/forum?id=FkDZLpK1Ml2}.

\bibitem[Vahdat et~al.(2021)Vahdat, Kreis, and Kautz]{vahdat2021score}
Vahdat, A., Kreis, K., and Kautz, J.
\newblock Score-based generative modeling in latent space.
\newblock \emph{Advances in Neural Information Processing Systems}, 34:\penalty0 11287--11302, 2021.

\bibitem[van~den Oord et~al.(2016)van~den Oord, Dieleman, Zen, Simonyan, Vinyals, Graves, Kalchbrenner, Senior, and Kavukcuoglu]{oord2016wavenet}
van~den Oord, A., Dieleman, S., Zen, H., Simonyan, K., Vinyals, O., Graves, A., Kalchbrenner, N., Senior, A., and Kavukcuoglu, K.
\newblock Wave{N}et: A generative model for raw audio.
\newblock \emph{arXiv preprint arXiv:1609.03499}, 2016.

\bibitem[Weininger(1988)]{weininger1988smiles}
Weininger, D.
\newblock Smiles, a chemical language and information system. 1. introduction to methodology and encoding rules.
\newblock \emph{Journal of chemical information and computer sciences}, 28\penalty0 (1):\penalty0 31--36, 1988.

\bibitem[Winter et~al.(2022)Winter, Bertolini, Le, Noe, and Clevert]{winter2022unsupervised}
Winter, R., Bertolini, M., Le, T., Noe, F., and Clevert, D.-A.
\newblock Unsupervised learning of group invariant and equivariant representations.
\newblock In Oh, A.~H., Agarwal, A., Belgrave, D., and Cho, K. (eds.), \emph{Advances in Neural Information Processing Systems}, 2022.
\newblock URL \url{https://openreview.net/forum?id=47lpv23LDPr}.

\bibitem[Wu et~al.(2022)Wu, Gong, Liu, Ye, and qiang liu]{wu2022diffusionbased}
Wu, L., Gong, C., Liu, X., Ye, M., and qiang liu.
\newblock Diffusion-based molecule generation with informative prior bridges.
\newblock In Oh, A.~H., Agarwal, A., Belgrave, D., and Cho, K. (eds.), \emph{Advances in Neural Information Processing Systems}, 2022.
\newblock URL \url{https://openreview.net/forum?id=TJUNtiZiTKE}.

\bibitem[Xu et~al.(2022)Xu, Yu, Song, Shi, Ermon, and Tang]{xu2022geodiff}
Xu, M., Yu, L., Song, Y., Shi, C., Ermon, S., and Tang, J.
\newblock Geodiff: A geometric diffusion model for molecular conformation generation.
\newblock \emph{arXiv preprint arXiv:2203.02923}, 2022.

\bibitem[Xu et~al.(2023)Xu, Powers, Dror, Ermon, and Leskovec]{xu2023geometric}
Xu, M., Powers, A., Dror, R., Ermon, S., and Leskovec, J.
\newblock Geometric latent diffusion models for 3d molecule generation.
\newblock In \emph{International Conference on Machine Learning}. PMLR, 2023.

\bibitem[Yu et~al.(2022)Yu, Li, Koh, Zhang, Pang, Qin, Ku, Xu, Baldridge, and Wu]{yu2022vectorquantized}
Yu, J., Li, X., Koh, J.~Y., Zhang, H., Pang, R., Qin, J., Ku, A., Xu, Y., Baldridge, J., and Wu, Y.
\newblock Vector-quantized image modeling with improved {VQGAN}.
\newblock In \emph{International Conference on Learning Representations}, 2022.
\newblock URL \url{https://openreview.net/forum?id=pfNyExj7z2}.

\bibitem[Zang \& Wang(2020)Zang and Wang]{zang2020moflow}
Zang, C. and Wang, F.
\newblock Moflow: an invertible flow model for generating molecular graphs.
\newblock In \emph{Proceedings of the 26th ACM SIGKDD International Conference on Knowledge Discovery \& Data Mining}, pp.\  617--626, 2020.

\bibitem[Zeng et~al.(2022)Zeng, Vahdat, Williams, Gojcic, Litany, Fidler, and Kreis]{zeng2022lion}
Zeng, X., Vahdat, A., Williams, F., Gojcic, Z., Litany, O., Fidler, S., and Kreis, K.
\newblock {LION}: Latent point diffusion models for 3d shape generation.
\newblock In Oh, A.~H., Agarwal, A., Belgrave, D., and Cho, K. (eds.), \emph{Advances in Neural Information Processing Systems}, 2022.
\newblock URL \url{https://openreview.net/forum?id=tHK5ntjp-5K}.

\end{thebibliography}
\bibliographystyle{icml2025}

\newpage
\appendix
\onecolumn
\section{Proof of the diffusion model}
\label{app:prop-diffusion}
We provide proofs for the derivation of several properties of the diffusion process in our model. For detailed explanation and discussion see \cite{ho2020denoising}.
\subsection{Marginal distribution of the diffusion process}
In the diffusion process, we have the marginal distribution of the data at any arbitrary time step $t$ in a closed form:
\begin{equation}
    \label{eq: q of any arbitrary time step}
q\left(\mathcal{G}_{t} \mid \mathcal{G}_{0}\right)=\mathcal{N}\left(\mathcal{G}_{t} ; \sqrt{\bar{\alpha}_{t}} \mathcal{G}_{0},\left(1-\bar{\alpha}_{t}\right) I\right).
\end{equation}
Recall the posterior $q\left(\mathcal{G}_{t} \mid \mathcal{G}_{0}\right)$ in Eq.2 (main document), we can obtain $\mathcal{G}_{t}$ using the reparameterization trick. A property of the Gaussian distribution is that if we add $\mathcal{N}(0,  \sigma^2_1I)$ and $\mathcal{N}(0, \sigma^2_2I)$, the new distribution is $\mathcal{N}(0,  (\sigma^2_1+\sigma^2_2)I)$
\begin{equation}
\label{eq: derivation of maginal distribution}
    \begin{split}
        \mathcal{G}_{t} &= \sqrt{\alpha_{t}}\mathcal{G}_{t-1} + \sqrt{1-\alpha_{t}}\epsilon_{t-1}\\
        &=\sqrt{\alpha_{t}\alpha_{t-1}}\mathcal{G}_{t-2} + \sqrt{\alpha_{t}(1-\alpha_{t-1})}\epsilon_{t-2} + \sqrt{1-\alpha_{t}}\epsilon_{t-1}\\
        &=\sqrt{\alpha_{t}\alpha_{t-1}}\mathcal{G}_{t-2} + \sqrt{1-\alpha_t\alpha_{t-1}}\bar{\epsilon}_{t-2}\\
        &=\dots\\
        &=\sqrt{\bar{\alpha}_t}\mathcal{G}_{0} + \sqrt{1-\bar{\alpha_t}}\bar{\epsilon},
    \end{split}
\end{equation}
where $\alpha_t=1-\beta_t$, $\epsilon$ and $\hat{\epsilon}$ are sampled from independent standard Gaussian distributions.
\subsection{The parameterized mean $\mathbf{\mu}_{\theta}$}
A learned Gaussian transitions $p_{\theta}\left(\mathcal{G}_{t-1} \mid  \mathcal{G}_{t}\right)$ is devised to approximate the $q\left(\mathcal{G}_{t-1} \mid  \mathcal{G}_{t}\right)$ of every time step: $p_{\theta}\left(\mathcal{G}_{t-1} \mid  \mathcal{G}_{t}\right)=\mathcal{N}\left(\mathcal{G}_{t-1} ; \boldsymbol{\mu}_{\theta}\left(\mathcal{G}_{t}, t\right), \sigma_{t}^{2} I\right)$. $\boldsymbol{\mu}_\theta$ is parameterized as follows:
\begin{equation}
    \boldsymbol{\mu}_{\theta}\left(\mathcal{G}_{t}, t\right)=\frac{1}{\sqrt{\alpha_{t}}}\left(\mathcal{G}_{t}-\frac{\beta_{t}}{\sqrt{1-\bar{\alpha}_{t}}} \boldsymbol{\epsilon}_{\theta}\left(\mathcal{G}_{t}, t\right)\right).
\end{equation}
 The distribution $q\left(\mathcal{G}_{t-1} \mid  \mathcal{G}_{t}\right)$ can be expanded by Bayes' rule:
\begin{equation}
\begin{split}
    q\left(\mathcal{G}_{t-1} \mid  \mathcal{G}_{t}\right) &= q\left(\mathcal{G}_{t-1} \mid  \mathcal{G}_{t}, \mathcal{G}_{0}, \right)\\
    &=q\left(\mathcal{G}_{t} \mid \mathcal{G}_{t-1}, \mathcal{G}_{0}\right) \frac{q\left(\mathcal{G}_{t-1} \mid \mathcal{G}_{0}\right)}{q\left(\mathcal{G}_{t} \mid \mathcal{G}_{0}\right)} \\
    &=q\left(\mathcal{G}_{t} \mid \mathcal{G}_{t-1}\right) \frac{q\left(\mathcal{G}_{t-1} \mathcal{G}_{0}\right)}{q\left(\mathcal{G}_{t} \mid \mathcal{G}_{0}\right)} \\
    &\propto \exp \left(-\frac{1}{2}\left(\frac{\left(\mathcal{G}_{t}-\sqrt{\alpha_{t}} \mathcal{G}_{t-1}\right)^{2}}{\beta_{t}}+\frac{\left(\mathcal{G}_{t-1}-\sqrt{\alpha_{t-1}} \mathcal{G}_{0}\right)^{2}}{1-\bar{\alpha}_{t-1}}-\frac{\left(\mathcal{G}_{t}-\sqrt{\alpha_{t}} \mathcal{G}_{0}\right)^{2}}{1-\bar{\alpha}_{t}}\right)\right) \\
    &=\exp \left(-\frac{1}{2}\left(\left(\frac{\alpha_{t}}{\beta_{t}}+\frac{1}{1-\bar{\alpha}_{t-1}}\right) \mathcal{G}_{t-1}^{2}-\left(\frac{2 \sqrt{\alpha_{t}}}{\beta_{t}} \mathcal{G}_{t}+\frac{2 \sqrt{\alpha_{t-1}}}{1-\bar{\alpha}_{t-1}} \mathcal{G}_{0}\right) \mathcal{G}_{t-1}+C\left(\mathcal{G}_{t}, \mathcal{G}_{0}\right)\right)\right)\\
    &\propto \exp (-\mathcal{G}_{t-1}^{2} + (\frac{\sqrt{\alpha_{t}}\left(1-\bar{\alpha}_{t-1}\right)}{1-\bar{\alpha}_{t}}\mathcal{G}_{t}+\frac{\sqrt{\bar{\alpha}_{t-1}} \beta_{t}}{1-\bar{\alpha}_{t}}\mathcal{G}_{0})\mathcal{G}_{t-1}),
\end{split}
\end{equation}
where $C\left(\mathcal{G}_{t}, \mathcal{G}_{0}\right)$ is a constant. We can find that $q\left(\mathcal{G}_{t-1} \mid  \mathcal{G}_{t}\right)$ is also a Gaussian distribution. We assume that:
\begin{equation}
    q\left(\mathcal{G}_{t-1} \mid \mathcal{G}_{t}, \mathcal{G}_{0}\right)=\mathcal{N}\left(\mathcal{G}_{t-1} ; \tilde{\boldsymbol{\mu}}\left(\mathcal{G}_{t}, \mathcal{G}_{0}\right), \tilde{\beta}_{t} I\right),
\end{equation}
where $\tilde{\beta}_{t}=1 /\left(\frac{\alpha_{t}}{\beta_{t}}+\frac{1}{1-\bar{\alpha}_{t-1}}\right)=\frac{1-\bar{\alpha}_{t-1}}{1-\bar{\alpha}_{t}} \cdot \beta_{t}$ and $\tilde{\boldsymbol{\mu}}_{t}\left(\mathcal{G}_{t}, \mathcal{G}_{0}\right)=\left(\frac{\sqrt{\alpha_{t}}}{\beta_{t}} \mathcal{G}_{t}+\frac{\sqrt{\bar{\alpha}_{t-1}}}{1-\bar{\alpha}_{t-1}} \mathcal{G}_{0}\right) /\left(\frac{\alpha_{t}}{\beta_{t}}+\frac{1}{1-\bar{\alpha}_{t-1}}\right)=\frac{\sqrt{\alpha_{t}}\left(1-\bar{\alpha}_{t-1}\right)}{1-\bar{\alpha}_{t}} \mathcal{G}_{t}+\frac{\sqrt{\bar{\alpha}_{t-1}} \beta_{t}}{1-\bar{\alpha}_{t}} \mathcal{G}_{0}$.
From Eq. \ref{eq: derivation of maginal distribution}, we have $\mathcal{G}_{t}==\sqrt{\bar{\alpha}_t}\mathcal{G}_{0} + \sqrt{1-\bar{\alpha_t}}\bar{\epsilon}$. We take this into $\tilde{\boldsymbol{\mu}}$:
\begin{equation}
    \begin{aligned}
\tilde{\boldsymbol{\mu}}_{t} &=\frac{\sqrt{\alpha_{t}}\left(1-\bar{\alpha}_{t-1}\right)}{1-\bar{\alpha}_{t}} \mathbf{x}_{t}+\frac{\sqrt{\bar{\alpha}_{t-1}} \beta_{t}}{1-\bar{\alpha}_{t}} \frac{1}{\sqrt{\bar{\alpha}_{t}}}\left(\mathbf{x}_{t}-\sqrt{1-\bar{\alpha}_{t}} \mathbf{\epsilon}_{t}\right) \\
&=\frac{1}{\sqrt{\alpha_{t}}}\left(\mathbf{x}_{t}-\frac{\beta_{t}}{\sqrt{1-\bar{\alpha}_{t}}} \mathbf{\epsilon}_{t}\right).
\end{aligned}
\end{equation}
$\boldsymbol{\mu}_\theta$ is designed to model $\tilde{\boldsymbol{\mu}}$. Therefore, $\boldsymbol{\mu}_\theta$ has the same formulation as $\tilde{\boldsymbol{\mu}}$ but parameterizes $\epsilon$:
\begin{equation}
    \boldsymbol{\mu}_{\theta}\left(\mathcal{G}_{t}, t\right)=\frac{1}{\sqrt{\alpha_{t}}}\left(\mathcal{G}_{t}-\frac{\beta_{t}}{\sqrt{1-\bar{\alpha}_{t}}} \boldsymbol{\epsilon}_{\theta}\left(\mathcal{G}_{t}, t\right)\right).
\end{equation}
\subsection{The ELBO objective}
It is hard to directly calculate log likelihood of the data. Instead, we can derive its ELBO objective for optimizing.  
\begin{equation}
\begin{split}
    \mathbb{E}\left[-\log p_{\theta}\left(\mathcal{G}\right)\right]
    &=-\mathbb{E}_{q\left(\mathcal{G}_{0}\right)} \log \left(\int p_{\theta}\left(\mathcal{G}_{0: T}, z_v\right) d \mathcal{G}_{1: T}\right) \\
    &=-\mathbb{E}_{q\left(\mathcal{G}_{0}\right)} \log \left(\int q\left(\mathcal{G}_{1: T} \mid \mathcal{G}_{0}\right) \frac{p_{\theta}\left(\mathcal{G}_{0: T}, z_v\right)}{q\left(\mathcal{G}_{1: T} \mid \mathcal{G}_{0}\right)} d \mathcal{G}_{1: T}\right) \\
    &=-\mathbb{E}_{q\left(\mathcal{G}_{0}\right)} \log \left(\mathbb{E}_{q\left(\mathcal{G}_{1: T} \mid \mathcal{G}_{0}\right)} \frac{p_{\theta}\left(\mathcal{G}_{0: T}, z_v\right)}{q\left(\mathcal{G}_{1: T} \mid \mathcal{G}_{0}\right)}\right) \\
    &\leq-\mathbb{E}_{q\left(\mathcal{G}_{0: T}\right)} \log \frac{p_{\theta}\left(\mathcal{G}_{0: T}, z_v\right)}{q\left(\mathcal{G}_{1: T} \mid \mathcal{G}_{0}\right)} \\
    &=\mathbb{E}_{q\left(\mathcal{G}_{0: T}\right)}\left[\log \frac{q\left(\mathcal{G}_{1: T} \mid \mathcal{G}_{0}\right)}{p_{\theta}\left(\mathcal{G}_{0: T}, z_v\right)}\right].
\end{split}
\end{equation}
Then we further derive the ELBO objective:
\begin{equation}
    \begin{split}
        \mathbb{E}_{q\left(\mathcal{G}_{0: T}\right)}\left[\log \frac{q\left(\mathcal{G}_{1: T} \mid \mathcal{G}_{0}\right)}{p_{\theta}\left(\mathcal{G}_{0: T}, z_v\right)}\right]
&=\mathbb{E}_{q}\left[\log \frac{\prod_{t=1}^{T} q\left(\mathcal{G}_{t} \mid \mathcal{G}_{t-1}\right)}{p_{\theta}\left(\mathcal{G}_{T}, z_v\right) \prod_{t=1}^{T} p_{\theta}\left(\mathcal{G}_{t-1} \mid \mathcal{G}_{t}, z_v\right)}\right]\\
&=\mathbb{E}_{q}\left[-\log p_{\theta}\left(\mathcal{G}_{T}, z_v\right)+\sum_{t=1}^{T} \log \frac{q\left(\mathcal{G}_{t} \mid \mathcal{G}_{t-1}\right)}{p_{\theta}\left(\mathcal{G}_{t-1} \mid \mathcal{G}_{t}, z_v\right)}\right]\\
&=\mathbb{E}_{q}\left[-\log p_{\theta}\left(\mathcal{G}_{T}, z_v\right)+\sum_{t=2}^{T} \log \frac{q\left(\mathcal{G}_{t} \mid \mathcal{G}_{t-1}\right)}{p_{\theta}\left(\mathcal{G}_{t-1} \mid \mathcal{G}_{t}, z_v\right)}+\log \frac{q\left(\mathcal{G}_{1} \mid \mathcal{G}_{0}\right)}{p_{\theta}\left(\mathcal{G}_{0} \mid \mathcal{G}_{1}, z_v\right)}\right]\\
&=\mathbb{E}_{q}\left[-\log p_{\theta}\left(\mathcal{G}_{T}, z_v\right)+\sum_{t=2}^{T} \log \left(\frac{q\left(\mathcal{G}_{t-1} \mid \mathcal{G}_{t}, \mathcal{G}_{0}\right)}{p_{\theta}\left(\mathcal{G}_{t-1} \mid \mathcal{G}_{t}, z_v\right)} \cdot \frac{q\left(\mathcal{G}_{t} \mid \mathcal{G}_{0}\right)}{q\left(\mathcal{G}_{t-1} \mid \mathcal{G}_{0}\right)}\right)+\log \frac{q\left(\mathcal{G}_{1} \mid \mathcal{G}_{0}\right)}{p_{\theta}\left(\mathcal{G}_{0} \mid \mathcal{G}_{1}, z_v\right)}\right]\\
&=\mathbb{E}_{q}\left[-\log p_{\theta}\left(\mathcal{G}_{T}, z_v\right)+\sum_{t=2}^{T} \log \frac{q\left(\mathcal{G}_{t-1} \mid \mathcal{G}_{t}, \mathcal{G}_{0}\right)}{p_{\theta}\left(\mathcal{G}_{t-1} \mid \mathcal{G}_{t}, z_v\right)}+\sum_{t=2}^{T} \log \frac{q\left(\mathcal{G}_{t} \mid \mathcal{G}_{0}\right)}{q\left(\mathcal{G}_{t-1} \mid \mathcal{G}_{0}\right)}+\log \frac{q\left(\mathcal{G}_{1} \mid \mathcal{G}_{0}\right)}{p_{\theta}\left(\mathcal{G}_{0} \mid \mathcal{G}_{1}, z_v\right)}\right]\\
&=\mathbb{E}_{q}\left[-\log p_{\theta}\left(\mathcal{G}_{T}, z_v\right)+\sum_{t=2}^{T} \log \frac{q\left(\mathcal{G}_{t-1} \mid \mathcal{G}_{t}, \mathcal{G}_{0}\right)}{p_{\theta}\left(\mathcal{G}_{t-1} \mid \mathcal{G}_{t}, z_v\right)}+\log \frac{q\left(\mathcal{G}_{T} \mid \mathcal{G}_{0}\right)}{q\left(\mathcal{G}_{1} \mid \mathcal{G}_{0}\right)}+\log \frac{q\left(\mathcal{G}_{1} \mid \mathcal{G}_{0}\right)}{p_{\theta}\left(\mathcal{G}_{0} \mid \mathcal{G}_{1}, z_v\right)}\right]\\
&=\mathbb{E}_{q}\left[\log \frac{q\left(\mathcal{G}_{T} \mid \mathcal{G}_{0}\right)}{p_{\theta}\left(\mathcal{G}_{T}, z_v\right)}+\sum_{t=2}^{T} \log \frac{q\left(\mathcal{G}_{t-1} \mid \mathcal{G}_{t}, \mathcal{G}_{0}\right)}{p_{\theta}\left(\mathcal{G}_{t-1} \mid \mathcal{G}_{t}, z_v\right)}-\log p_{\theta}\left(\mathcal{G}_{0} \mid \mathcal{G}_{1}, z_v\right)\right]\\
&=\mathbb{E}_{q}\Bigg{[}\underbrace{D_{\mathrm{KL}}\left(q\left(\mathcal{G}_{T} \mid \mathcal{G}_{0}\right) \| p_{\theta}\left(\mathcal{G}_{T}, z_v\right)\right)}_{L_{T}}+\\
&\sum_{t=2}^{T} \underbrace{D_{\mathrm{KL}}\left(q\left(\mathcal{G}_{t-1} \mid \mathcal{G}_{t}, \mathcal{G}_{0}\right) \| p_{\theta}\left(\mathcal{G}_{t-1} \mid \mathcal{G}_{t}, z_v\right)\right)}_{L_{t}}\underbrace{-\log p_{\theta}\left(\mathcal{G}_{0} \mid \mathcal{G}_{1}, z_v\right)}_{L_{0}}\bigg{]}.
    \end{split}
\end{equation}

\section{Explanation of Proposition \ref{MVAE} }
\label{app:prop-ae}
Consider a geometric graph $\mathcal{G} = \langle \mathbf{x}, \mathbf{h} \rangle$, encoder $\mathcal{E}$ and decoder $\mathcal{D}$, such that $\mathcal{G} = \mathcal{D}(\mathcal{E}(\mathcal{G}))$. Then we assume that the action transformation is $g$, and through $g$, $\mathcal{G}$ is transformed from the SE(3) group to $\hat{\mathcal{G}} = T_{g}\mathcal{G} = \langle \mathbf{h},\mathbf{R}\mathbf{x}+\mathbf{t} \rangle$, and input it into the autoencoder. Since the encoding function is unchanged, we can deduce $\mathcal{E}(\mathcal{G})=\mathcal{E}(\hat{\mathcal{G}})$, so the geometric shape reconstructed by the decoder is still $\mathcal{G} = \mathcal{D}(\mathcal{E}(\hat{\mathcal{G}}))$, instead of $\hat{\mathcal{G}}$. This makes it impossible to calculate the reconstruction error based on $\mathcal{G}$ and $\hat{\mathcal{G}}$. The solution is to add a function $\psi$ to extract the group action $g$. Then send it to the decoder, we can apply the group action to the generated $\mathcal{G}$ to recover $\hat{\mathcal{G}}$, solving the above problem.

\section{MVAE Architecture Details}
\label{app:MVAE detail}
In the MVAE we proposed, they are all parameterized with EGNN\cite{satorras2021enflow} as the backbone. EGNN is a type of graph neural network that satisfies the equivariance properties in \ref{eq:equivariance:ae} and \ref{Equivariant Markov prop}. In EGNN, the molecular geometry is usually regarded as a point cloud structure without specifying the connecting bonds between atoms. However, in practice, we use the interatomic distance 2\text{\AA} as the close distance criterion to construct close covalent bonds and long-range van der Waals forces. The point cloud structure symbol is $\mathcal{G}$, and the interactions between all atoms $\mathrm{v}_{i} \in \mathcal{V}$ are modeled. Each node $\mathrm{v}_{i}$ is embedded with coordinates $\mathrm{x}_{i} \in \mathbb{R}^3$ and atomic features $\mathrm{h}_{i} \in \mathbb{R}^d$. Then EGNN consists of multiple equivariant convolutional layers $\mathbf{x}^{l+1},\mathbf{h}^{l+1},\mathbf{v}^{l+1} = \mathrm{EGCL}[\mathbf{x}^l,\mathrm{h}^l,\mathrm{v}^l]$, and each single layer is defined as:\\
\begin{equation}
\begin{aligned}
\label{app:EGNN}
    \mathbf{m}_{ij} &= \phi_{e}\left(\mathrm{h}_{i}^l,\mathrm{h}_{j}^l,d_{ij}^2,a_{ij}\right),\\
    \mathrm{h}_{i}^{l+1}&=\phi_{h}(\mathrm{h}_{i}^l,\sum_{j \neq i}\tilde{e}_{ij}),\\
    \mathrm{v}_{i}^{l+1} &= \phi_{v}(\mathrm{h}_i^l)\mathrm{v}_i^l+C\sum_{j \neq i}\left(\mathrm{x}_i^l - \mathrm{x}_j^l\right)\phi_{x}(\mathrm{m}_{ij}),\\
    \mathrm{x}_{i}^{l+1} &= \mathrm{x}_{i}^{l}+\mathrm{v}_i^{l+1},
\end{aligned}
\end{equation}
where $l$ represents the layer index. $\tilde{e}_{ij} = \phi_{inf}(\mathbf{m}_{ij})$ serves as the attention weight to re-weight the information passed from different edges. $d_{ij}=\mid\mid\mathrm{x}_{i}^{l}-\mathrm{x}_{j}^{l} \mid\mid$ represents the pairwise distance between atoms $\mathrm{v}_{i}$ and $\mathrm{v}_{j}$, and $a_{ij}$ is an optional edge feature. We also incorporate velocity features in each layer. We generally set the initial velocity $\mathrm{v}_{i}^{(0)}$ to 0, and we update the position $\mathrm{x}_{i}^{l+1}$ by the velocity $\mathrm{v}_{i}^{l+1}$. The velocity is updated by the function $\phi_{v}:\mathbb{R}^{N} \rightarrow \mathbb{R}^{1}$, and the function is equivariant. We give the proof below. First, we prove that the velocity update, that is, the third line of the formula, maintains equivariance, that is, we want to prove:
\[
Q\mathrm{v}_i^{l+1} = \phi_{v}(\mathrm{h}_i^l)Q\mathrm{v}_i^{init}+C\sum_{j \neq i}(Q\mathrm{x}_i^l+g-[Q\mathrm{x}_j^l+g])\phi_x(m_{ij})
\]\\
\textit{Derivation.}\\
\begin{equation}
\begin{aligned}
     Q\mathrm{v}_i^{l+1} &= \phi_{v}(\mathrm{h}_i^l)Q\mathrm{v}_i^{init}+C\sum_{j \neq i}(Q\mathrm{x}_i^l+g-[Q\mathrm{x}_j^l+g])\phi_x(m_{ij})\\
     &=Q\phi_v(\mathrm{h}_i^l)\mathrm{v}_i^{iniit}+QC\sum_{j \neq i}(\mathrm{x}_i^l-\mathrm{x}_j^l)\phi_x(\mathrm{m}_{ij})\\
     &=Q(\phi_v(\mathrm{h}_i^l)\mathrm{v}_i^{iniit}+C\sum_{j \neq i}
     (\mathrm{x}_i^l-\mathrm{x}_j^l)\phi_x(\mathrm{m}_{ij}))\\
     &=Q\mathrm{v}_i^{l+1}
    \end{aligned}
\end{equation}
Finally, it is straightforward to show the second equation is also equivariant, that is we want to show $Q\mathrm{x}_i^{l+1}+g=Q\mathrm{x}_i^{l}+g+Q\mathrm{v}_i^{l+1}$:\\
\textit{Derivation.}\\
\begin{equation}
    \begin{aligned}
    Q\mathrm{x}_i^{l}+g+Q\mathrm{v}_i^{l+1} &= Q(\mathrm{x}_i^{l}+\mathrm{v}_i^{l+1})\\
    &=Q\mathrm{x}_i^{l+1}+g
\end{aligned}
\end{equation}
The above proof ensures that the E(n) transformation on our input point cloud will output the same transformation, so that $\mathrm{h}^{l+1},Q\mathrm{x}^{l+1}+g,Q\mathrm{v}^{l+1}=\mathrm{EGCL}[\mathrm{h}^l,Q\mathrm{x}^l+g,Q\mathrm{v}^{init},\mathcal{E}]$ holds.

\section{Equivariant Markov Kernels Details}
\label{app:Markov Kernels}
First, keep the invariance of feature $\mathrm{h}$. We use $\mathrm{Edge MLP}$ to obtain edge embedding as follows:
\begin{equation}
\mathrm{h}_{e_{ij}}=\mathrm{MLP}(\mathrm{d}_{ij},e_{ij})
\end{equation}

Where $\mathrm{d}_{ij}=\mid\mid \mathrm{x}_{i} - \mathrm{x}_{j} \mid\mid_2$ represents the Euclidean distance between the coordinates of atom $i$ and atom $j$, and $e_{ij}$ represents the edge feature between the $i$th atom and the $j$th atom. We use $a$ to represent the atomic feature, and then use the $L$ layer of \textbf{Schnet}\cite{schutt2017schnet} to achieve the invariance effect:
\begin{equation}
\begin{aligned}
     \mathrm{h}_{i,\mathrm{a}}^0 &= \mathrm{MLP}(\mathrm{a}_{i}),\mathrm{h}_{i,\mathrm{x}}^0=\mathrm{MLP}(\mathrm{x}_{i}),h_{i}^0=[\mathrm{h}_{i,\mathrm{a}}^0,\mathrm{h}_{i,\mathrm{x}}],\\
    h_i^{l+1} &= \sigma\left(W_{0}^l\mathrm{h}_i^{l}+\sum_{j \in N(i)}W_1^l\phi_{w}(\mathrm{d}_{ij})\odot W_2^l\mathrm{h}_j^l\right),
\end{aligned}
\end{equation}

Where $l \in (0,1,\dots,L)$ represents the $l$th layer of $Schnet$, $W^l$ represents the learned weights. Then, we regard $\mathrm{h}_i$ output by $Schnet$ as node embedding, $\sigma(\cdot)$ represents a nonlinear activation function, such as $\mathrm{ReLU}$, and $\phi_w(\cdot)$ represents the weight network. Then, the final output of $Schnet$ is represented as a node embedding.

To estimate the gradient of the log density of atomic features, we use a layer of \textbf{Node MLP} to map the latent vector output by $schnet$ to a score vector.
\begin{equation}
    s_\theta(\mathrm{a}_i)=\mathrm{MLP}(\mathrm{h}_i)
\end{equation}

The equivariance of the coordinate $\mathrm{x}$ in 3D space is achieved, decomposed into pairwise distances, and the product of the learned edge information and the end node vector of the same edge is connected into a \textbf{Pairwise Block}, which is then input into the \textbf{Distance MLP} to obtain the distance gradient between pairs of atoms.
\begin{equation}
    s_\theta(\mathrm{d}_{ij})=\mathrm{MLP}([\mathrm{h}_i,\mathrm{h}_j,\mathrm{h}_{e_{ij}}])
\end{equation}
here, we omit the time coordinate $t$ of $s_\theta$ and only perform the analysis on a single time step for simplicity.

Then, the \textbf{Dist-transition Block} is used for transformation to integrate the information of the pairwise distance and the fractional vector of the atomic coordinates $\mathrm{x}$, as shown below:
\begin{equation}
    s_\theta(\mathrm{x}_i)=\sum_{j \in N(i)}\frac{1}{\mathrm{d}_{ij}}\cdot s_\theta(\mathrm{d}_{ij})\cdot (\mathrm{x}_i - \mathrm{x}_j)
\end{equation}
where $s_\theta(\mathrm{x}_i)$ is invariant to translation because it only depends on the symmetry invariant element $\mathrm{d}$, and $\mathrm{x}_i-\mathrm{x}_j$ is rotationally and translationally equivariant, so $s_\theta(\mathrm{x}_i)$ is equivariant.

We proceed to prove that our diffusion model composed of Markov Kernels is equivariant. Following previous work \cite{xu2022geodiff,hoogeboom2022equivariant}, we will omit the trivial scalar feature $\mathrm{a}$ and focus on analyzing the latent variable $\mathrm{Z}$. The proof shows that when the initial distribution $p(\mathrm{z}^T)$ is invariant and the transfer distribution $p(\mathrm{z}_{\mathrm{x}}^{t-1} \mid \mathrm{z}_{\mathrm{x}}^{t})$ is equivariant, then the marginal distribution $p(\mathrm{z}_{\mathrm{x}}^{T})$ will be time-invariant, in particular, including $p(\mathrm{z}_\mathrm{x}^0)$. Similarly, since the decoder $(EGNN)$ output distribution $p(\mathrm{x}\mid \mathrm{z}_{\mathrm{x}}^0)$ is equivariant, we can obtain that our final data distribution $p(\mathrm{x})$ is unchanged everywhere.

\textit{Proof.}The justification formally can be derived as follow:

\textbf{Condition:} We know that $p(\mathrm{z}_{\mathrm{x}}^T)=\mathcal{N}(0,I)$ is invariant under rotation, \textit{e.g.}, $p(\mathrm{z}_{\mathrm{x}}^T)=p(R\mathrm{z}_{\mathrm{x}}^T)$.

\textbf{Derivation:} For $t \in {1,\dots, T}$, let $p(\mathrm{z}_{\mathrm{x}}^{t-1}\mid\mathrm{z}_{\mathrm{x}}^{t})$ be an equivariant distribution, \textit{i.e.}, $p(\mathrm{z}_{\mathrm{x}}^{t-1}\mid\mathrm{z}_{\mathrm{x}}^{t})=p(R\mathrm{z}_{\mathrm{x}}^{t-1}\mid R\mathrm{z}_{\mathrm{x}}^{t})$ for all orthogonal $R$. Assuming $p(\mathrm{z}_{\mathrm{x}}^t)$ is an invariant distribution, \textit{i.e.}, $p(\mathrm{z}_{\mathrm{x}}^t)=p(R\mathrm{z}_{\mathrm{x}}^t)$ for all $R$, then we have:

\begin{equation}
    \begin{aligned}
    p(\mathrm{Rz}_{\mathrm{x}}^{t-1}) &=\int_{\mathrm{z}_{\mathrm{x}}^t}p(\mathrm{R}\mathrm{z}_{\mathrm{x}}^{t-1}\mid\mathrm{z}_{\mathrm{x}}^t)p(\mathrm{z}_{\mathrm{x}}^t) &\text{Chain Rule}\\
    &=\int_{\mathrm{z}_{\mathrm{x}}^t}p(\mathrm{R}\mathrm{z}_{\mathrm{x}}^{t-1}\mid\mathrm{RR^{-1}}\mathrm{z}_{\mathrm{x}}^t)p(\mathrm{RR^{-1}}\mathrm{z}_{\mathrm{x}}^t) &\text{Multiply by $\mathrm{RR^{-1}}=\mathrm{I}$}\\
    &=\int_{\mathrm{z}_{\mathrm{x}}^t}p(\mathrm{R}\mathrm{z}_{\mathrm{x}}^{t-1}\mid\mathrm{R^{-1}}\mathrm{z}_{\mathrm{x}}^t)p(\mathrm{R^{-1}}\mathrm{z}_{\mathrm{x}}^t)  &\text{Equivariance \& Invariance} \\
    &=\int_{\mathrm{z}_{\mathrm{x}}^t}p(\mathrm{z}_{\mathrm{x}}^{t-1}\mid \mathrm{y})p(\mathrm{y})\cdot \underbrace{\det\mathrm{R}}_{=1} &\text{change of Variables $\mathrm{y}=\mathrm{R^{-1}z_{x}^t}$}\\
    &=p(\mathrm{z_x^{t-1}}),
    \end{aligned}
\end{equation}
Therefore, $p(\mathrm{z_x^{t-1}})$ is invariant. By induction, $p(\mathrm{z_x^{T-1}}),\dots,p(\mathrm{z_x^{0}})$ is invariant. In addition, since the decoder $p(\mathrm{x}\mid\mathrm{z_x^0})$ is also equivariant. By the same derivation, we can also conclude that the distribution of our generated molecules $p(\mathrm{x})$ is also invariant.

\section{Dataset}
\label{app:dataset}
\textbf{QM9} QM9 dataset contains over 130K small molecules with quantum chemical properties which each consist of up to 9
heavy atoms or 29 atoms including hydrogens. On average, each molecule contains 18 atoms. For a fair comparison, we follow
the previous work \cite{anderson2019cormorant} to split the data into training, validation and test set, which each partition
contains 100K, 18K and 13K molecules respectively.\\
LMDM is trained by Adam \cite{DBLP:journals/corr/KingmaB14} optimizer for 200K iterations (about 512 epochs) with a batch size of 256
and a learning rate of 0.001.

\textbf{Geom} Following previous work \cite{hoogeboom2022equivariant}, we evaluate MDM on a larger scale dataset GEOM \cite{axelrod2022geom}. Compared to QM9, the size of molecules in GEOM is much larger, in which is up to 181 atoms and
46 atoms on average (including hydrogens). We obtain the lowest energy conformation for each molecule, and finally we have
290K samples for training.\\
LMDM is trained by Adam \cite{DBLP:journals/corr/KingmaB14} optimizer for 200K iterations (about 170 epochs) with a batch size of 256 and a learning rate of 0.001.

\section{Ablation Studies}
\label{app:exp-ablation}
In this section, we provide additional experimental results on QM9 to demonstrate the effect of the model design. We perform ablation experiments on two variables, latent dimension $k$ and regularization square, and the results are reported in Table \ref{Tab:3}.
\begin{table*}[!ht]
\centering
\caption{Results of ablation study with different model designs. Metrics are calculated with 10000 samples generated from each setting.}
\label{Tab:3}
\begin{threeparttable}
\begin{tabular}{lcccc}\\\toprule 
& \multicolumn{4}{c}{QM9}\\
~ &  Validity $(\%)$ $\uparrow$ &  Uniqueness $(\%)$ $\uparrow$ &  Novelty $(\%)$ $\uparrow$ &  Stability $(\%)$ $\uparrow$ \\\midrule
LMDM($k=2$,\textit{KL})*& 86.3 & 95.9 & 79.3 & 82.1\\
LMDM($k=2$,\textit{ES}) & 97.7 & 93.5 & 87.9 & 89.2\\
LMDM($k=1$,\textit{KL}) & 87.4 & 93.7 & 78.5 & 85.4\\
\rowcolor{lightgray}
LMDM($k=1$,\textit{ES}) & \textbf{98.8} & \textbf{95.2} & \textbf{92.1} &\textbf{90.8}\\
\bottomrule
\end{tabular}
\begin{tablenotes}
\small
\item *Note that this reported result is already the best result we achieved for \textit{KL}.
\end{tablenotes}
\end{threeparttable}
\end{table*}

We first discuss different regularization methods for autoencoders, \textit{i.e.}, \textit{KL-reg} and \textit{ES-reg}, where the invariant feature dimension of the latent variable is fixed to 1. Following previous work \cite{rombach2022high}, for \textit{KL-reg}, we use a weight parameter of 1 in our experiments. However, we observe unexpected failures and extremely poor performance in our experiments. As shown in Table \ref{Tab:3}, the performance of all models with \textit{KL} is very different from that of models with \textit{ES}. The models with \textit{KL} terms are also extremely unstable during training, often making numerical errors and causing model training failures. We observe in the generated molecules in our experiments that equivariant latent features often tend to converge to highly dispersed means and very small variances, which may be the cause of numerical problems in the \textit{KL} term calculation. We therefore turn to using \textit{ES} to constrain the encoder by early stopping the encoder training, which can limit the numerical range of the latent features. We will learn more about the other effects of KL regularization.

From Table 3, we also observe that LMDM performs better when $k=1$. This also shows that lower dimensions can reduce the complexity of generative modeling and facilitate training LMDM. The performance of LMDM on QM9 is very similar when $k$ is set to 1 or 2. In practice, we set $k$ to 1 for the QM9 dataset, and set $k$ to 2 for the GEOM-Drug dataset, which contains more atoms.

\section{Visualization Results}
\begin{figure}[!ht]
    \centering
    \includegraphics[width=1.0\linewidth]{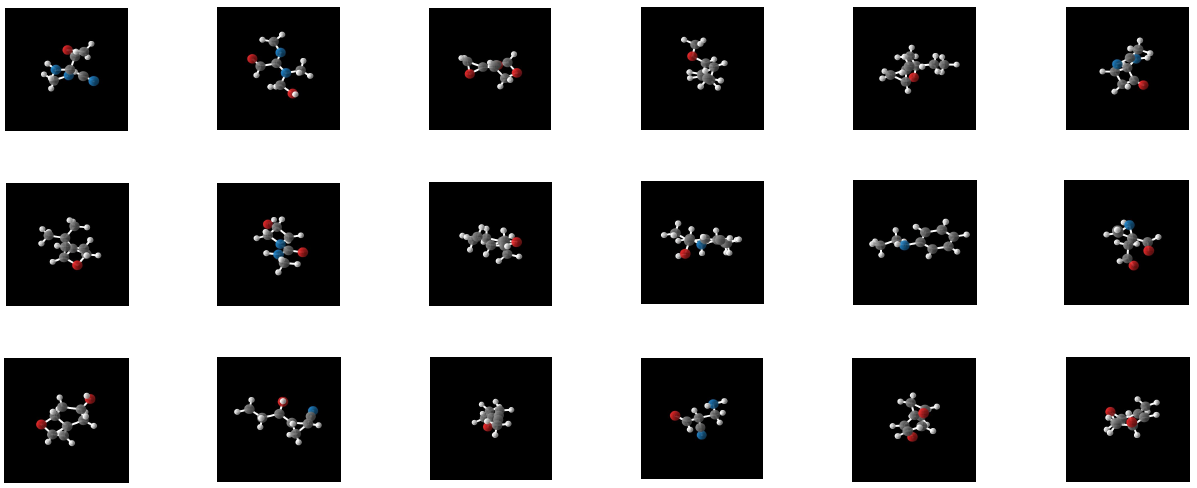}
    \vspace{-15pt}
    \caption{Molecules generated from LMDM trained on QM9.}
    \label{app:fig-qm9}
\end{figure}

\begin{figure}[!ht]
    \centering
    \includegraphics[width=1.0\linewidth]{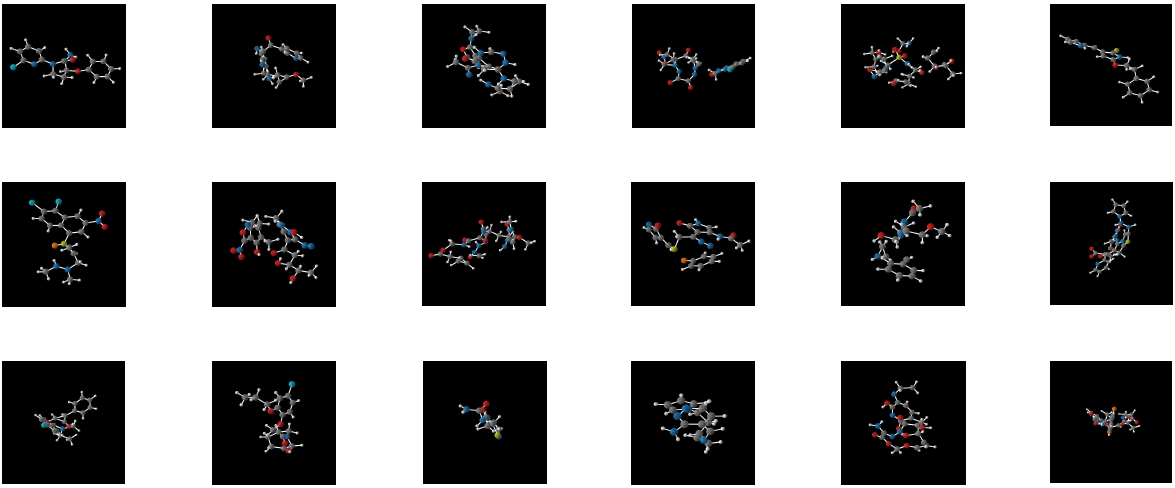}
    \vspace{-15pt}
    \caption{Molecules generated from LMDM trained on GEOM-Drug.}
    \label{app:fig-drug}
\end{figure}

In this section, we show the visualization of molecules generated by LMDM. As shown in Figure\ref{app:fig-qm9} and Figure\ref{app:fig-drug}, they are samples generated by sampling from the models trained on the QM9 and GEOM-Drug datasets, respectively. These samples are generated by random sampling, and we can see that the perspective of observing the molecular structure may not be perfect, but this can to some extent reflect the diversity of the spatial structure of the generated molecules, because it extends as much as possible in all directions of the zero center of mass (COM)\cite{kohler2020equivariant} space in the random process. In the figure \ref{app:fig-drug}, we can see that there are two GEOM-Drug molecules composed of small molecules. This phenomenon is usually caused by the instability of the molecular spatial structure due to the large molecular weight, but it is not actually a problem. After all, the molecular-level stability of the original dataset is only 86.5\%, and it is very common in non-autoregressive molecular generation models\cite{zang2020moflow,jo2022score}. Removing the small molecular weight components can achieve a repair effect\cite{xu2023geometric}.

\end{document}